\pdfoutput=1
\documentclass[11pt]{article}

\usepackage[]{EMNLP2023}

\usepackage{times}
\usepackage{latexsym}

\usepackage[T1]{fontenc}
\usepackage[utf8]{inputenc}

\usepackage{microtype}

\usepackage{graphicx}
\usepackage{amsmath}
\usepackage{amsthm}
\usepackage{booktabs}
\usepackage{subfigure}
\usepackage{amssymb}
\usepackage{multirow}
\usepackage{microtype}
\usepackage{float}
\usepackage{booktabs}
\usepackage{soul}
\usepackage{url}
\usepackage{color}
\usepackage{makecell}
\usepackage{xcolor}
\usepackage[linesnumbered,ruled,vlined]{algorithm2e}
\usepackage{inconsolata}

\title{How to Enhance Causal Discrimination of Utterances:\\ A Case on Affective Reasoning}

\author{Hang Chen\and Xinyu Yang \and Jing Luo \\
Xi'an Jiaotong University \\ \texttt{\{albert2123,luojingl\}@stu.xjtu.edu.cn}\\
\texttt{yxyphd@mail.xjtu.edu.cn}
        \And 
        Wenjing Zhu\\ Du Xiao Man Inc.\\ \texttt{zhuwenjing02@duxiaoman.com}}
  \SetKwInput{KwInput}{Input}                % Set the Input
  \SetKwInput{KwOutput}{Output} 

\begin{document}
\maketitle
\begin{abstract}
  Our investigation into the Affective Reasoning in Conversation 
  (ARC) task highlights the challenge of causal discrimination. 
  Almost all existing models, 
  including large language models (LLMs), 
  excel at capturing semantic correlations 
  within utterance embeddings but fall short in determining the  
  specific causal relationships. 
  To overcome this limitation, 
  we propose the incorporation of \textit{i.i.d.} noise terms 
  into the conversation process, thereby constructing a 
  structural causal model (SCM). It explores   
  how distinct causal relationships of fitted embeddings can be 
  discerned through independent conditions. 
  To facilitate the implementation of deep learning, 
  we introduce the \textit{cogn} frameworks to 
  handle unstructured conversation data, 
  and employ an autoencoder architecture to regard 
  the unobservable noise as learnable ``implicit causes.'' 
  Moreover, 
  we curate a synthetic dataset that includes \textit{i.i.d.} noise. 
  Through comprehensive experiments, 
  we validate the effectiveness and interpretability of 
  our approach. Our code is available in \url{https://github.com/Zodiark-ch/mater-of-our-EMNLP2023-paper}. 
\end{abstract}

\section{Introduction}

Nowadays, numerous conversation recognition tasks 
(such as Emotion Recognition in Conversation (ERC) task
~\cite{pereira2022deep,thakur2023audio}, Intent Recognition (IR) task
~\cite{ye2023msmix,ni2023attention} and Dialogue Act Recognition (DAR) task
~\cite{arora2023joint}) 
have shown promising performance in specialized supervised 
and unsupervised methods. 
Considering the RoBERTa pre-trained model~\cite{liu2019roberta} 
as the examples, 
``My eyelids are fighting'' and ``I want to sleep,'' 
which have similar semantics but different tokens 
can be well fitted within embeddings. (i.e., 
these two embeddings exhibit a strong resemblance 
via certain metrics such as cosine similarity.) 

However, when it comes to the relationship between two utterances, 
denoted as $A$ and $B$, wherein their embeddings can be fitted, 
various possible relationships exist: $A$ acts as the cause of $B$ 
($A \rightarrow B$), $A$ acts as the outcome of $B$ 
($A \leftarrow B$), or more complex, $A$ and $B$ are both influenced 
by a common cause ($A \leftarrow C \rightarrow B$), and so on. 
Particularly in reasoning tasks 
~\cite{uymaz2022vector,fengemowoz}, 
it is crucial for these methods to transcend the mere fitting 
of embeddings and possess the capacity to discriminate 
diverse causal relationships.  
(i.e., the ability of \textbf{causal discrimination})
~\cite{bao2022multi,shirai2023towards}. 

To specifically investigate the causal discrimination capability 
of existing methods in conversation, 
we narrow down our research to a particular task: 
Affective Reasoning in Conversation (ARC), which has included 
Emotion-Cause Pair Extraction (ECPE)~\citet{xia-ding-2019-emotion} and  
Emotion-Cause Span Recognition (ECSR)~\citet{poria2021recognizing}. 

We begin with conducting tests to evaluate the causal discrimination 
of existing methods including the large language models (LLMs)
~\cite{kasneci2023chatgpt}. One typical evaluation involves 
the causal reversal test: for emotion-cause utterance pairs with true labels 
($A$, $B$) representing a causal relationship of $B \rightarrow A$, 
we scrutinize the predictions generated by the existing methods 
using both positive pairs ($A$, $B$) and negative pairs ($B$, $A$). 
The results reveal that all the examined methods performed similarly 
across the two sample types. As we are concerned, 
they lacked causal discriminability. 
(Details are shown in Section~\ref{seccar})

In order to discriminate different causal relationships 
between two similar embeddings, we construct the 
dialogue process as a Structural Causal Model (SCM).  
Many endeavors 
~\citep{cheng2022evaluation,nogueira2022methods} 
supporting that \textit{i.i.d.} noise of SCM 
could facilitate the discrimination of causal relationships 
when fitting two variables. 
Under the presence of noise, 
each utterance is not only explicitly influenced by the other 
utterances but also implicitly influenced by 
the \textit{i.i.d.} exogenous noise. 
Consequently, this framework ensures that two fitted 
embeddings result in diverse causal relationships, 
which are determined by corresponding independent conditions 
between the residual terms and embeddings. 
For simplicity, we refer to other utterances as explicit causes 
and exogenous noise as implicit causes.  

Furthermore, to enable the learnability of such causal discrimination 
within embeddings, we propose a common skeleton, named 
\textit{centering one graph node (cogn)} skeleton for each 
utterance derived from some broadly accepted prior hypotheses. 
It can address the challenges arising from 
variable-length and unstructured dialogue samples. 
Subsequently, we develop an autoencoder architecture 
to learn the unobservable implicit causes. 
Specifically, we consider the implicit causes as 
latent variables and utilize a graph attention network (GAT)
~\cite{velivckovic2017graph} to encode its representation. 
Additionally, the decoder leverages the inverse matrix 
of the causal strength, ensuring an accurate retrieval 
of the causal relationships. 

Finally, we conduct extensive experimental evaluations: 
1) our approach significantly outperforms existing methods including  
prominent LLMs (GPT-3.5 and GPT-4) in two affective reasoning tasks 
(ECPE and ECSR) and one emotion recognition task (ERC), 
demonstrating its effectiveness in affective reasoning. 
2) our method exhibits a significant reduction  
in false predictions for negative samples across 
three causal discrimination scenarios. 
3) we curate a synthetic dataset with implicit causes 
to visualize the latent variable in our implementation.

Our contribution is four-fold: 

\begin{itemize}
  \item We formulated the dialogue process as an 
  SCM and analyzed the causal relationships represented by 
  different independent conditions. 
  \item We devised the \textit{cogn} skeleton to address 
  the problems of variable-length and unstructured dialogue samples.
  \item We adopted an autoencoder architecture to overcome 
  the unobservability of implicit causes and make it learnable. 
  \item We constructed a synthetic dataset with implicit causes 
  and conducted extensive evaluations of our proposed method. 
\end{itemize}

\section{Related Works and Challenges}
\subsection{Task Definition}
For notational consistency, we use the following terminology.  
The \textbf{target utterance} $U_{t}$ is the $t^{th}$ utterances of 
a conversation $\mathcal{D}=(U_{1},U_{2},U_{3},\dots, U_{N})$ 
where $N$ is the maximum number of utterances in this conversation and 
$0<t\leqslant N$. 
The \textbf{emotion label} $Emo_{t}$ denotes the emotion type of $U_{t}$. 
The \textbf{emotion-cause pair (ECP)} is a pair $(U_{t},U_{i})$, 
where $U_{i}$ is the $i^{th}$ utterance of this conversation. In the 
ECP, $U_{t}$ represents the emotion utterance and $U_{i}$ is the 
corresponding cause utterance. 
Moreover, the \textbf{cause label} $C_{t,i}$ denotes 
the cause span type of the ECP $(U_{t},U_{i})$.

Thus, in a given text, \textbf{ERC} is the task of 
identifying all $Emo_{t}$. Moreover, \textbf{ECPE} aims to 
extract a set of ECPs and \textbf{ECSR} aims to identify 
all $C_{t,i}$. 

\subsection{Affective Reasoning in Conversation}
\citet{chen2018emotionlines} introduced the pioneering work on ERC 
due to the growing availability of public conversational data. 
Building upon this,~\citet{xia-ding-2019-emotion} 
further advanced the field by proposing the ECPE that jointly 
identifies both emotions and their corresponding causes. Moreover,
~\citet{poria2021recognizing} has extended ECPE into conversations 
and proposed a novel ECSR task, specifically designed to 
identify ECP spans within conversation contexts. 
More recently, increasing works have indicated the 
crucial role played by accurate inference models 
in facilitating complex reasoning within these tasks, such as the 
assumption about interlocutors~\citep{zhang2019modeling,lian2021decn,shen-etal-2021-directed} 
and context~\citep{ghosal-etal-2019-dialoguegcn,Shen2021DialogXLAX,chen2022learning}. 

\subsection{Challenge of Affective Reasoning}\label{seccar}
We examined the performance of a range of methods 
for addressing affective reasoning in conversations, 
including both unsupervised approaches 
(large language models (LLMs), BERT-based pre-trained models) 
and supervised approaches (task-related approaches). 

Overall, all the methods demonstrated a lack of discriminability 
on two types of challenges: 
\begin{itemize}
  \item Samples where emotional utterances and causal utterances 
  are interchanged. For a dialogue instance, if the 
  ECP is $(U_{1}, U_{2})$ 
  ($U_{2}$ is the cause of $U_{1}$), the prediction results obtained 
  by the existing methods tend to include both 
  $(U_{1}, U_{2})$ and $(U_{2}, U_{1})$.
  \item Samples with indirect connections. For example, 
  if the ECPs in a conversation are 
  $(U_{1}, U_{2})$ and $(U_{2}, U_{3})$, 
  the prediction results obtained by the methods often include an 
  additional pair $(U_{1}, U_{3})$.
\end{itemize}
\begin{table}
  \footnotesize
  \centering
  \resizebox{\linewidth}{!}{
  \begin{tabular}{|c|c|c|c|c|c|}
    \hline
    \multirow{2}{*}{Methods} & \multicolumn{2}{c}{Challenge 1} \vline&  \multicolumn{3}{c}{Challenge 2} \vline\\
    \cline{2-6}
    &($U_{a}$,$U_{b}$)&($U_{b}$,$U_{a}$)&($U_{a}$,$U_{b}$)&($U_{b}$,$U_{c}$)&($U_{a}$,$U_{c}$)\\
    \hline
     GPT-3.5&112 & 102& 108&114 & 109\\
     GPT-4&127 &114 &111 & 105& 103\\
    \hline
    RoBERTa&95 & 97&94 &91 & 83\\
    RoBERTa$^{+}$&97 &91& 105& 101&106\\
    \hline
    RANK-CP&142 & 125& 147& 129& 131\\
     ECPE-2D&151 &153 &142 &138 & 146\\
     EGAT& 166& 154&157 & 139&148 \\
     \hline 
  \end{tabular}}
  \caption{For the two challenges mentioned in Section~\ref{seccar}, 
  we conducted tests on a subset of 200 samples from the RECCON 
  dataset. We recorded the number of samples identified 
  by above methods. In the second row of Challenge 1, 
  we showed the count of samples where these methods extracted 
  the negative pairs in reverse  cause order. 
  Similarly, in the third row of Challenge 2, we showed the 
  count of samples where these methods extracted negative indirect pairs.}
  \label{tabchallenge}
\end{table}
We evaluated the performance of existing methods 
on these two challenges, and the detailed results are shown 
in Table~\ref{tabchallenge}. All evaluated methods extracted 
a nearly equal number of negative samples as positive samples. 
Considering their performance 
in broad research domains, both unsupervised and supervised methods 
could demonstrate a desirable fitting ability to 
capture the semantic similarity between two utterances. 
This often apparently results in satisfactory performance 
in most tasks. However, when it comes to more specific 
causal relationships within semantically similar sentences 
(such as reasoning tasks), 
they may not exhibit the same level of ``intelligence'' and 
output some ``pseudo-correlation''.

In the area of causal discovery, 
Causal Markov and Faithfulness Assumptions
~\citep{spirtes2000constructing,colombo2012learning,ogarrio2016hybrid}, 
provide insights into capturing more specific causal relationships 
in the situation of the above challenges. 
Considering two similar variables: $A$ and $B$ that can be fitted, 
the independence tests enable the determination of 
specific causal relationships, such as 
``$A \rightarrow B$,'' ``$B \rightarrow A$,'' or 
``$A \rightarrow L \rightarrow B$''. 
More recently, the Structural Causal Model (SCM) 
~\citep{shimizu2006linear,shimizu2014bayesian,sanchez2019estimating}
built upon the independent noise assumptions 
has emerged as an effective approach to the limitation of 
Markov equivalence classes in distinguishing causal relationships. 
The noise terms (also called exogenous variables) for each variable, 
enables methods such as Independent Component Analysis (ICA) 
to identify more comprehensive causal relationships 
between the two fitted variables.

\section{Methodology}
In this section, we begin by outlining 
incorporating \textit{i.i.d.} noise terms into a dialogue model 
to construct an SCM in Section~\ref{secscm}, 
demonstrating independent residual allowing for the identification of 
more specific causal relations within pairs of fitted utterances. 
Next, to mitigate conflicts between SCM models and 
dialogue data, we designed \textit{cogn} skeletons with 
six instantiations in Section~\ref{seccse}. 
Finally, we propose a deep learning 
implementation to tackle the issue of noise 
being unknown in dialogue data in Section~\ref{secai}. 

\subsection{Structural Causal Model}\label{secscm}
\begin{figure}
  \includegraphics[width=1\linewidth]{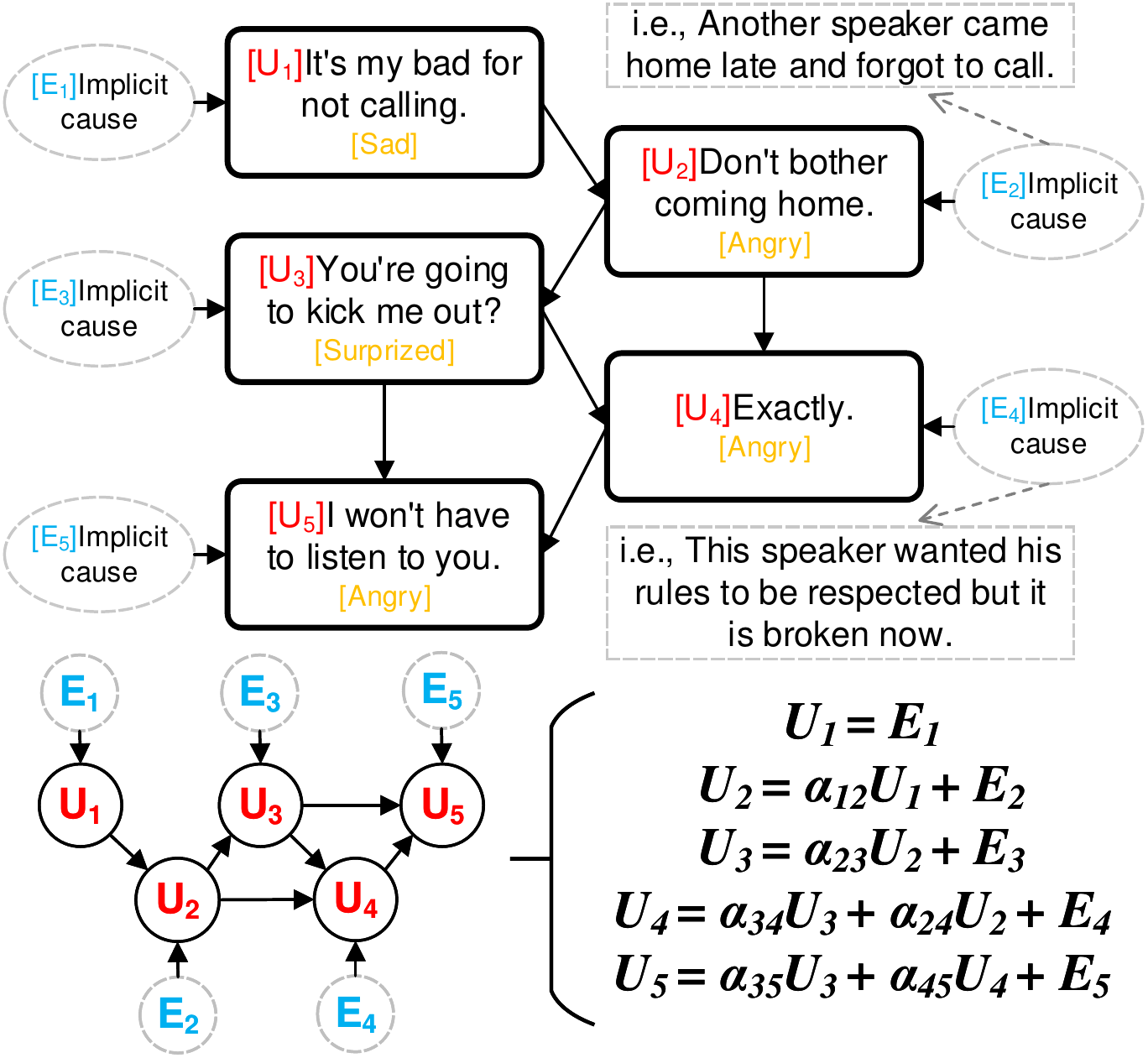}
  \caption{The conversation case with five utterances. In the SCM, 
  we assume that each utterance $U_{i}$ has a corresponding 
  implicit cause $E_{i}$, and has several explicit causes. 
  i.e., $U_{4}$ has an implicit cause $E_{4}$ and two explicit causes 
  $U_{3}$ and $U_{2}$. In the lower part of the figure, SCM adopts 
  $U_{t}=\sum \alpha_{it}U_{i}+E_{t}$ 
  to denote these relationships and formalize the conversation as a DAG. }
  \label{figscm}
\end{figure}
In order to imbue causal discriminability into the fitting process 
of two relevant utterances, we algebraically construct the 
conversation model as a Structural Causal Model (SCM). 

\textbf{Definition 1}: An SCM of a dialogue 
is a 3 tuple $\langle U, E, \mathcal{F}\rangle $, where 
$U$ is the set of utterances 
$U=\{U_{i}\}^{N}_{i=1}$, $E$ is the set of exogenous noises 
$ E=\{E_{i}\}^{N}_{i=1}$ corresponding to each $U_{i}$, 
$N$ is the number of utterances. Note that each $E_{i}$ is 
independent in the SCM. Structural equations 
$\mathcal{F}=\{f_{i}\}^{N}_{i=1}$ are functions that determine 
  $U$ with $U_{i}=f_{i}(rel_{U_{{i}}})+E_{i}$, 
  where $rel_{U_{{t}}}$ denotes a set of utterances 
  that point to the $U_{t}$. 

Definition 1 establishes the construction of a novel 
computational model for dialogue process, as exemplified in 
Figure~\ref{figscm}. In such a computational model, 
each utterance is endogenous and influenced by an 
independent exogenous variable. For simplicity, 
we refer to the variable $U$ as the explicit causes and 
the variable $E$ as the implicit causes. 
The independence of the implicit causes makes the residual terms 
meaningful during the fitting of any two utterances. 

\textbf{Definition 2}: The relationship of two utterances 
$X$ and $Y$ in a dialogue 
is causal discriminable, from the independent conditions: 
\begin{itemize}
  \item $\Sigma_{X} \perp \!\!\! \perp Y, 
  \Sigma_{Y} \not \! \perp \!\!\! \perp X
  \Rightarrow Y \rightarrow X$
  \item $\Sigma_{X} \not \! \perp \!\!\! \perp Y, 
  \Sigma_{Y} \perp \!\!\! \perp X
  \Rightarrow X \rightarrow Y$
  \item $\Sigma_{X} \not \! \perp \!\!\! \perp Y, 
  \Sigma_{Y} \not \! \perp \!\!\! \perp X
  \Rightarrow L \rightarrow X, L \rightarrow Y$
  \item $\Sigma_{X} \perp \!\!\! \perp Y, 
  \Sigma_{Y} \perp \!\!\! \perp X
  \Rightarrow X \rightarrow L, Y \rightarrow L$
\end{itemize}
where $\Sigma$ represents the residual terms in fitting process. 
(The proof is shown in Appendix~\ref{pod2}.) 

\textbf{Example 1}: A 4-utterance dialogue SCM 
$D=\{\{U_{1}, U_{2}, U_{3}, U_{4}\}, 
\{E_{1}, E_{2}, E_{3}, E_{4}\}, \{a,b,c\}\}$ with the true relationships 
are $U_{1}=E_{1}$, $U_{2}=aU_{1}+E_{2}$, $U_{3}=bU_{1}+E_{3}$, 
$U_{4}=cU_{3}+E_{4}$. The fitting of $U_{2}$ 
with $U_{3}$ and $U_{4}$ yield 
$U_{2}=\frac{a}{b}U_{3}+0U_{4}+\Sigma_{U_{2}} $, 
while the fitting of $U_{3}$ with $U_{2}$ and $U_{4}$ yield 
$U_{3}=\frac{b}{a}U_{2}+\frac{1}{c}U_{4}+\Sigma_{U_{3}} $. 
Additionally, 
the fitting of $U_{4}$ with $U_{2}$ and $U_{3}$ lead to 
$U_{4}=0U_{2}+cU_{3}+\Sigma_{U_{4}} $. 

In Example 1, it is observed that any two utterances 
can be fitted together as they are mutually dependent. 
However, causal discriminability can be employed to differentiate 
their distinct causal structures. For instance, 
the residual term $\Sigma_{U_{3}}$ is not independent of $U_{4}$, 
while $\Sigma_{U_{4}}$ is independent of $U_{3}$, 
indicating that $U_{3}$ is a cause of $U_{4}$. 
Furthermore, the residual term $\Sigma_{U_{3}}$ is not 
independent of $\Sigma_{U_{2}}$, 
and $\Sigma_{U_{2}}$ is not independent of $U_{3}$, 
implying the presence of common cause  ($U_{1}$) 
between $U_{2}$ and $U_{3}$. 

 \subsection{Causal Skeleton Estimation}\label{seccse}
Establishing a skeleton is the first step in causal discovery, 
as different skeletons provide distinct learning strategies 
for recovering the relationships between variables. 
However, utterances differ from the variables that causal discovery 
often uses. Specifically, each conversation has a different amount ($N$) of utterances, 
and different inter-utterances relationships related to the 
context. Hence, it is intractable to build a general causal skeleton 
with fixed nodes and edges to describe all conversation samples. 

Fortunately, several published GNN-based approaches
~\citep{shen-etal-2021-directed,ishiwatari-etal-2020-relation,ghosal-etal-2019-dialoguegcn,lian2021decn,zhang2019modeling} 
in ERC have proposed and verified a common hypothesis to settle down this issue. 
The Hypotheses are elaborated on in Appendix~\ref{has}. 

\begin{figure}[h]
  \includegraphics[width=1\linewidth]{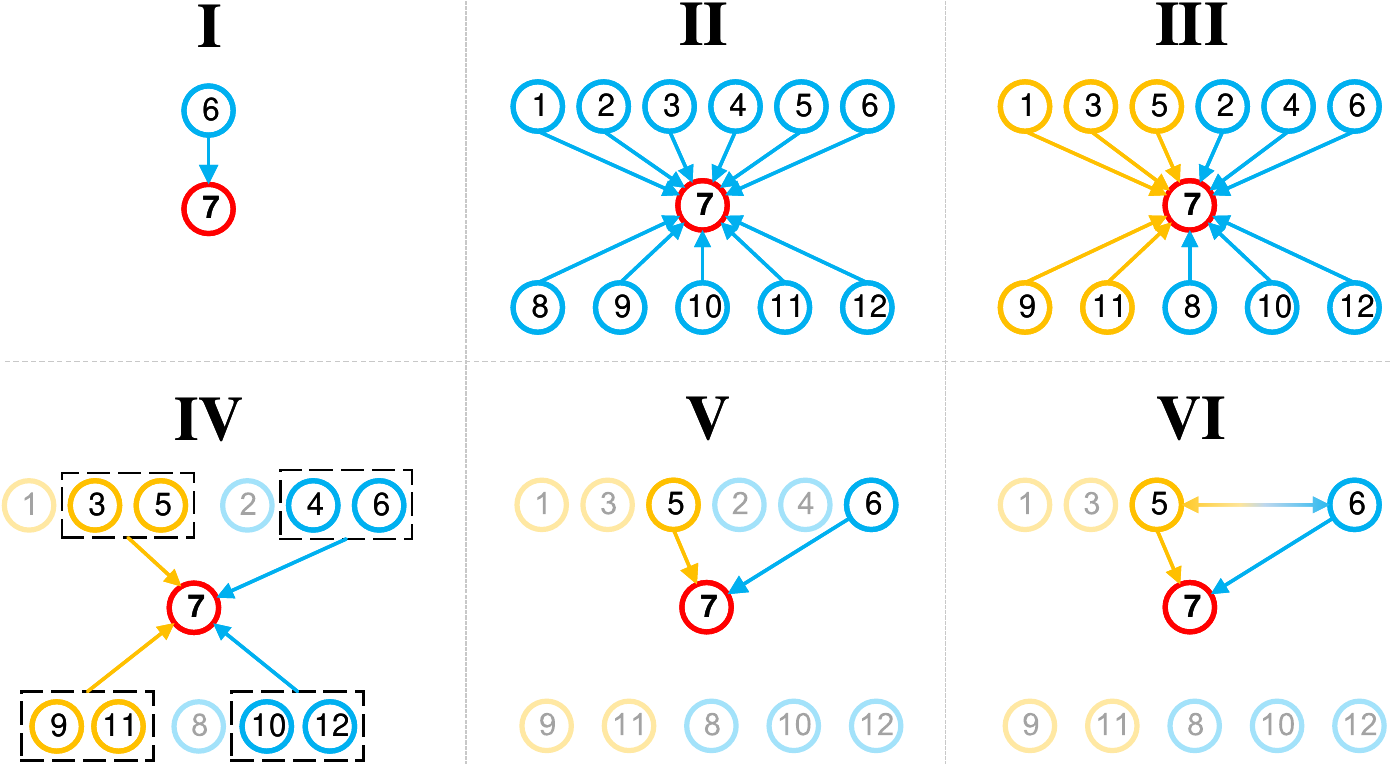}
  \caption{Six \textit{cogn} skeletons from a conversation case with 
  12 utterances. We adopted the $7$-th utterance as the target utterance (Red). 
  Orange nodes denote the utterances of the same speaker as the target utterance, 
  and blue ones denote those belonging to other speakers. 
  Arrow represents the information propagated from one utterance to 
  another, and the bi-way arrow represents the influence-agnostic 
  relationship. The black dash box represents the slide windows.}
  \label{figskeleton}
\end{figure}

Figure~\ref{figskeleton} showcases the design of 
six \textit{cogn} skeletons, derived from the latest works 
that have employed one or more of these hypotheses. 
The statistic and specific algorithms are also 
shown in Appendix~\ref{has}. Note that we only 
conduct experiments for $\textbf{II-VI}$ because our structure is hard to 
apply with the recurrent-based skeleton. 

\subsection{Approach Implementation}\label{secai}
\begin{figure*}
  \includegraphics[width=1\linewidth]{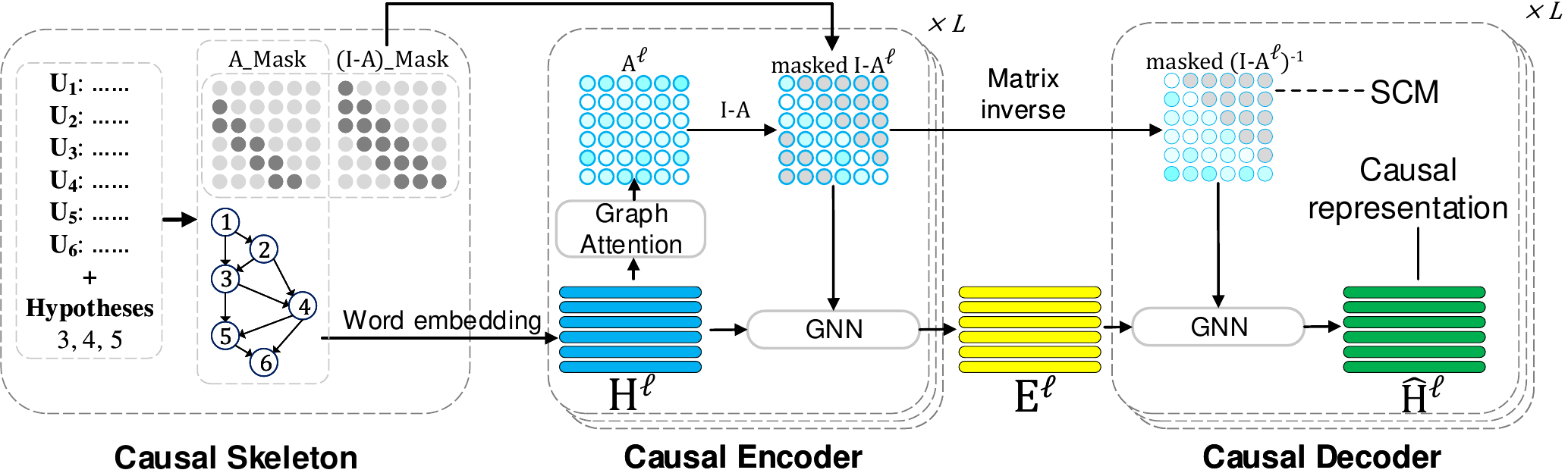}
  \caption{Processing of our approaches, with a six-utterances conversation case 
  as the input. Causal skeleton indicates which utterances (nodes) 
  are used for aggregation. For each layer $\ell$, we collect representations 
  $H^{\ell}$ for all utterances where each row represents one utterance. 
  Causal Encoder yields the implicit causes $E^{\ell}$, the input for Decoder learning 
  the causal representation. In all matrices, 
  light gray nodes represent the masked part.}
  \label{figCAE}
\end{figure*}
From a given causal skeleton, a linear SCM 
can be equivalently represented as: 
\begin{equation}
  U_{t}=\sum_{i\in rel_{t}} \alpha_{i,t}U_{i}+E_{t}  
  \label{eqn1}
\end{equation}
\noindent
where $rel_{t}$ denotes a set of utterances that point to the $U_{t}$ 
($7$-th utterance) in Figure~\ref{figskeleton}, 
$E_{t}$ represents the exogenous variable towards the variable $U_{t}$ in SCM, 
as well as the implicit cause towards the utterance $U_{t}$ in conversation. 
Furthermore, we denote the word embedding of $U$ by $H={h_{1}, h_{2}, \dots, h_{N}}$, 
and the relationships between utterances in rows can also be written as:
$H=A^{T}H+E$, where $A_{i,t}\neq0$ stands for a directed edge from $U_{i}$ to $U_{t}$ in SCM. 
Thus we can define the Graph $\mathcal{G}=(\mathcal{V}, \mathcal{E})$ 
with adjacency matrix $A_{i,i}=0$ for all $i$. 

However, in this equation, only $H$ is known. 
The unknown variable $A$ is typically the target of inference, 
while the unknown variable $E$ represents exogenous variables  
that implicitly influence each utterance, 
such as the speaker's memory, experiences, or desires. 
~\cite{krych2007think,sidera2018theory}
These factors are typically missing in existing conversation 
resources. Therefore, determining $A$ based on completely 
unknown $E$ is another problem we aim to address. 

Hence, we treat $A^{T}$ as an autoregression matrix of the $\mathcal{G}$, 
and then $E$ can be yielded by an auto-encoder model. The whole 
process reads: 
\begin{equation}
  E=f((I-A^{T})H) 
  \label{eqn3}
\end{equation}
\begin{equation}
  \widehat{H}=g((I-A^{T})^{-1}E)  
  \label{eqn4}
\end{equation}
where $f(\cdot)$ and $g(\cdot)$ represent the encoder and decoder neural networks respectively. 
Encoder aims to generate an implicit cause $E$, and Decoder 
devotes to yielding a causal representation $\widehat{H}$. 
From Equation~\ref{eqn1}, causal representation $\widehat{H}_{t}$
reasons about the fusion relations of heterogeneous 
explicit causes $\sum_{i\in rel_{t}}H_{i}$  and implicit cause $E_{t}$. 
The details of this process are shown in Figure~\ref{figCAE}. 

\textbf{Encoder.} We use the graph attention mechanism to learn the 
adjacency matrix $A$ and construct a hierarchical GNN to instantiate 
the $f(\cdot)$. $\ell={1, 2, \dots, L-1}$ represents the 
layer of GNN. Thus, for each utterance at the $\ell$-th layer, 
the $A^{\ell}_{i,t}$ computed by attention mechanism is a weighted combination 
of $h^{\ell}_{t}$ for each directly related utterance 
$U_{i} (i\in rel_{t})$: 
\begin{equation}
  A^{\ell}_{i,t}=\frac{LeakyReLU(e_{i,t}^{\ell})}{\sum_{j\in rel_{t}}LeakyReLU(e_{j,t}^{\ell})} 
\label{eqn5}
\end{equation}
\begin{equation}
  e_{i,t}^{\ell}=\overrightarrow{h}_{i}W^{\ell}_{i(row)}+(\overrightarrow{h}_{t}W^{\ell}_{t(col)})^{T}
  \label{eqn6}
\end{equation}
where $W^{\ell}_{row}\in \mathbb{R}^{N\times1}$ and $W^{\ell}_{col}\in \mathbb{R}^{N\times1}$ 
are the learnable parameters in the graph attention. Moreover, the 
GNN aggregates the information from the neighbor utterances as following: 
\begin{equation}
  H^{\ell+1}=eLU((I-(A^{\ell})^{T})H^{\ell}W^{\ell})
  \label{eqn7}
\end{equation}
where $W^{\ell}$ stands for parameters in the corresponding layer. 
From the final layer of the evaluation process, 
by extracting $A^{L-1}$ computed in Equation~\ref{eqn5}, 
the marginal or conditional ``distribution'' of $H$ is obtained, 
showing how to discover Causal Graph $\mathcal{G}$ from $\mathcal{D}$. 
Besides, by extracting $H^{L}$ in Equation~\ref{eqn7}, we can 
obtain the independent embedding for the implicit causes $E=MLP(H^{L})$. 

\textbf{Decoder.}
We utilize the $A$ and $E$ computed from Encoder to generate the 
causal representation $\widehat{H}$. With a fixed adjacency matrix $A$, 
the GNN aggregates the information of implicit causes 
from neighbor nodes as follows: 
\begin{equation}
  \widehat{E}^{\ell+1}=eLU((I-(A^{L})^{T})^{-1}E^{\ell}M^{\ell})
  \label{eqn9}
\end{equation}
where $M^{\ell}$ is parameters in the corresponding layer. As the same 
architecture as the encoder, $\widehat{H}=MLP(E^{L})$. Additionally, 
the plug-in RNN is integrated with GNN to address the appetite 
of \textbf{Hypothesis 6}: 
\begin{equation}
  \widehat{E}^{\ell+1}=GRU^{\ell}(\widehat{E}^{\ell},p^{\ell})
  \label{eqn10}
\end{equation}
where $p^{\ell}$ is the state of GRU model, 
with $p$ computed by self-attention proposed by~\citet{thost2021directed}.

\subsection{Optimization}\label{sec4.3}

In our approach, $\widehat{H}$ and $H$ acts identically 
under the \textbf{linear} SCM model. Similarly, 
$\widehat{H}$ should be aligned with $H$ in emotion dimensions 
under the \textbf{non-linear} SCM model. In short, 
we adopt an auxiliary loss measuring the Kullback-Leibler (KL) 
divergence~\cite{joyce2011kullback} of $\widehat{H}$ and $H$ 
mapped into the exact emotion dimensions. Moreover, implicit causes 
$E$ is one of the crucial influence factors  on $\widehat{H}$, 
so that the loss aims to impose the constraint that $\widehat{H}$ 
is the embedding of our need to ensure generating correct $E$. 
\begin{equation}
  Loss_{KL}=\sum_{t} \sum_{e\in Emo_{t}}p_{e}(\widehat{U_{t}}) \log \frac{p_{e}(\widehat{U_{t}})}{p_{e}(U_{t})}   
  \label{eqn11}
\end{equation}
where $e$ is any emotion type in $Emo_{t}$, 
$p_{e}$ denotes the probability 
labeled with emotion $e$. In the whole process of 
ARC tasks, we followed
~\cite{wei-etal-2020-effective,poria2021recognizing} 
to add several losses of ECPE and ECSR respectively. 

Furthermore, we would like to explain the difference between our 
approach and Variational Auto-Encoder (VAE)
~\citep{2014Auto}. 
The output of the encoder in VAE is $q_{\phi}(Z)$. 
With this estimation of $\widehat{q}_{\phi}(Z)$, 
we can measure the variation $\xi(q_{\phi}(Z))$ 
(also called $\nabla_{\phi}ELBO(\widehat{q}_{\phi}(Z))$) 
to obtain the approximation estimation of $ELBO(q)$. 
In contrast, our output is $E$, a fixed matrix 
rather than a distribution. In other words, 
the VAE depends on the prior distribution over  
the latent variables, whereas our approach 
has a dependency on the consistency of $H$ and $\widehat{H}$, 
which is non-sampling and non-distributive.

\section{Experiments}

In this section, we conduct extensive experiments to answer the 3 
research questions:

\textbf{RQ1:} How effective is our method in affective reasoning tasks? 

\textbf{RQ2:} How do we justify the causal discriminability of our method?

\textbf{RQ3:} How do we gauge the difference between the latent variable 
 $E$ and designed implicit causes?

\subsection{Datasets, Implementation and Baselines}

\begin{table}
  \footnotesize
  \centering
  \resizebox{\linewidth}{!}{
  \begin{tabular}{|c|c|c|c|c|c|c|}
    \hline
    \multirow{2}{*}{Dataset}&\multicolumn{3}{c}{conversations}\vline&\multicolumn{3}{c}{tasks}\vline\\
    \cline{2-7}
     &Train&Val&Test&ERC&ECPE&ECSR\\
    \hline
    DailyDialog & 11118 &1000&1000&$\surd$ &$-$ &$-$ \\
    MELD  & 1038 &114&280&$\surd$ &$-$ &$-$ \\
    EmoryNLP &713&99&85&$\surd$ &$-$ &$-$ \\
    IEMOCAP&100&20&31&$\surd$ &$-$ &$-$ \\
    RECCON-DD&833&47&225&$-$ &$\surd$ &$\surd$ \\
    RECCON-IE&$-$ &$-$ &16&$-$ &$\surd$ &$\surd$ \\
    Synthetic data&833&47&225&$-$ &$\surd$ &$\surd$ \\
    \hline
  \end{tabular}}
  \caption{ The statistics of seven datasets}
  \label{tabdataset}
\end{table}

We use six real datasets for three affective reasoning tasks and one 
synthetic dataset for justifying $E$ in our model. 
The statistics of them are shown in Table~\ref{tabdataset}. 
Appendix~\ref{D} depicts the detailed introductions of each dataset. 

We adopt the consistent benchmarks of the SOTA methods in three tasks, 
including the pre-training language model, hyper-parameters, $t$-tests,  
and metrics. The implementation details are shown in Appendix~\ref{ID}. 

According to the hypotheses of these baselines, for each \textit{cogn} skeleton, 
we choose one recent SOTA work:  
II: \textbf{DialogXL}~\citep{Shen2021DialogXLAX}. 
III: \textbf{EGAT}~\citep{chen2022learning}. 
IV: \textbf{RGAT}~\citep{ishiwatari-etal-2020-relation}. 
V: \textbf{DECN}~\citep{lian2021decn}. 
VI: \textbf{DAG-ERC}~\citep{shen-etal-2021-directed}. 

\subsection{Overall Performance (RQ1)}

Table~\ref{tabecpeecsr} reports the results of ECPE and ECSR, 
with $p<$0.01 in the \textit{t}-test,  
where the best improvement and best performance 
both concentrate on $\textbf{VI}$. With the visualization of Appendix~\ref{vcg}, 
we infer that the upper triangular adjacency matrix of DAG-ERC, not 
restricted by the backpropagation, benefits from \textbf{Hypothesis 6}. 
Moreover, $\textbf{II}$ lags farthest behind in the ECPE while achieving  
the second best in the ECSR, showing that the reliance on a hypothesis 
is not equal in different tasks.  
Furthermore, without \textbf{Hypotheses 1} and \textbf{6}, $\textbf{III}$, $\textbf{IV}$, and $\textbf{V}$ 
are far from the best performance since \textbf{Hypothesis 1} has the 
maximum identifying space, and \textbf{Hypothesis 6} supplies the highest number of  
information passing. Finally, it is worth noting that three skeleton-agnostic 
baselines and unsupervised methods perform poorly in 
the RECCON-IE dataset, indicating that our models have stronger 
representation learning capabilities as well as highlighting 
the continued research value of affective reasoning tasks.

\begin{table}
  \footnotesize
  \centering
  \resizebox{\linewidth}{!}{
  \begin{tabular}{|c|c|c|c|c|c|}
    \hline
    \multirow{2}{*}{Skt}&\multirow{2}{*}{model}&\multicolumn{2}{c}{ECPE in RECCON}\vline&\multicolumn{2}{c}{ECSR in RECCON}\vline\\
    \cline{3-6}
    & &DD($\pm\sigma_{10}$)&IE&DD($\pm\sigma_{10}$)&IE\\
    \hline
    \multirow{2}{*}{$-$}&GPT-3.5&38.13&39.55&10.49&5.36\\
    &GPT-4&46.29&49.32&16.81&18.39\\
    &RoBERTa&53.91$\pm$1.5&38.77&31.52$\pm$0.8&20.16\\
    &RoBERTa$^{+}$&54.62$\pm$1.1&38.26&33.28$\pm$0.7&26.37\\
    &RANK-CP$\dagger$&63.51$\pm$2.1&41.56&26.57$\pm$0.8&18.99\\
    &ECPE-2D$\dagger$&64.35$\pm$1.7&47.42&34.41$\pm$0.1&22.03\\
    \hline
    \multirow{2}{*}{$\textbf{II}$}&DialogXL$\dagger$&61.92$\pm$1.7&50.31&\textbf{35.79$\pm$0.5}&21.78\\
    &+Ours&\textbf{64.74$\pm$1.6}&\textbf{51.23}&34.63$\pm$0.2&\textbf{27.92}\\
    \hline
    \multirow{2}{*}{$\textbf{III}$}&EGAT&68.05$\pm$1.5&53.43&29.68$\pm$0.7&16.42\\
    &+Ours&\textbf{69.16$\pm$1.2}&\textbf{53.81}&\textbf{30.5$\pm$0.2}&\textbf{18.55}\\
    \hline
    \multirow{2}{*}{$\textbf{IV}$}&RGAT$\dagger$&69.02$\pm$1.9&52.48&\textbf{30.39$\pm$0.4}&17.49\\
    &+Ours&\textbf{70.12$\pm$2.1}&\textbf{53.93}&30.24$\pm$0.5&\textbf{19.31}\\
    \hline
    \multirow{2}{*}{$\textbf{V}$}&DECN$\dagger$&68.32$\pm$1.5&51.73&30.7$\pm$0.9&18.47\\
    &+Ours&\textbf{68.84$\pm$1.7}&\textbf{53.89}&\textbf{31.88$\pm$0.2}&\textbf{20.13}\\
    \hline
    \multirow{2}{*}{$\textbf{VI}$}&DAG-ERC$\dagger$&70.36$\pm$1.5&55.7&40.12$\pm$0.7&24.89\\
    &+Ours&\textcolor{red}{\textbf{73.17$\pm$1.1}}&\textcolor{red}{\textbf{56.67}}&\textcolor{red}{\textbf{42.14$\pm$0.1}}&\textcolor{red}{\textbf{30.41}}\\
    \hline
  \end{tabular}}
  \caption{Overall performance in ECPE and ECSR tasks. We additionally compare 
  four unsupervised approaches and two baselines not belonging to any skeleton: 
  RANK-CP~\cite{wei-etal-2020-effective}, ECPE-2D~\cite{ding-etal-2020-ecpe}. 
  The RoBERTa$^{+}$ represents the large RoBERTa version (1024 dimensions). The DD 
  and IE are two subsets (see Appendix~\ref{D}).}
  \label{tabecpeecsr}
\end{table}

We further conducted six sets of ablation 
experiments to study the effects of different modules. In Table~\ref{tabablation}, 
we summarized results under the following cases: replacing $Loss_{KL}$ 
with $BCE$ loss function ($BCE$); 
removing the $Loss_{KL}$ (w/o $Loss_{KL}$); 
replacing Decoder module with a Linear layer (w/o Decoder); 
removing the RNN module (w/o \textbf{Hypo 6}); 
adding the edges from successors to predecessors (w/o \textbf{Hypo 5}); 
reducing the speaker types to one (w/o \textbf{Hypo 4}). 

\begin{table}
  \footnotesize
  \centering
  \resizebox{\linewidth}{!}{
  \begin{tabular}{|c|c|c|c|c|c|}
    \hline
    \multirow{2}{*}{Model}&\multicolumn{5}{c}{Categories}\vline\\
    \cline{2-6}
    &$\textbf{II}$&$\textbf{III}$&$\textbf{IV}$&$\textbf{V}$&$\textbf{VI}$\\
    \hline
    Ours&64.74&69.16&70.12&68.84&73.17\\
    \hline
    $BCE$&$\downarrow$0.62&$\downarrow$0.04&$\downarrow$0.15&$\downarrow$0.16&$\downarrow$0.29\\
    w/o $Loss_{KL}$&$\downarrow$2.18&$\downarrow$1.95&\textbf{$\downarrow$2.42}&$\downarrow$1.33&$\downarrow$1.58\\
    w/o Decoder&\textbf{$\downarrow$3.59}&\textbf{$\downarrow$2.79}&$\downarrow$2.11&\textbf{$\downarrow$2.83}&\textbf{$\downarrow$4.14}\\
    \hline
    w/o Hypo 6&$-$&$-$&$-$&$-$&$\downarrow$1.59\\
    w/o Hypo 5&$-$&$-$&$-$&$\downarrow$2.34&$\downarrow$1.88\\
    w/o Hypo 4&$-$&\textbf{$\downarrow$3.67}&\textbf{$\downarrow$2.72}&\textbf{$\downarrow$3.15}&\textbf{$\downarrow$4.19}\\
    \hline
  \end{tabular}}
  \caption{Ablation results}
  \label{tabablation}
\end{table}

As shown in Table~\ref{tabablation}, $BCE loss$ performs similarly 
to $Loss_{KL}$; thus, we empirically demonstrate that our auxiliary loss 
is essentially different from $Loss_{KL}$ in VAE. The F1 score 
decreases heavily without auxiliary loss or decoder, these two 
are necessary ingredients for building complete processing to learn 
the causal representation via $E$. Besides, \textbf{Hypotheses 
4, 5}, and \textbf{6} are all critical but removing \textbf{Hypothesis 4} leads to 
the highlight degradation in 3 skeletons. This result corroborates the 
theory of~\citet{lian2021decn} and~\citet{shen-etal-2021-directed}, 
who state that speaker identity is the strong inductive bias in 
conversation. Finally, it is expected to see that skeleton 
with \textbf{Hypotheses 4, 5, and 6} should be the closest 
to perfection while the DAG-ERC$+$Ours indeed achieves the SOTA. 

Furthermore, Appendix~\ref{oear} reports the results of ERC task and 
sensitivity experiments to analyze how our model 
performs in different $L$ and $k$. 

\subsection{Relationship analysis (RQ2)}

\begin{table}
  \footnotesize
  \centering
  \resizebox{\linewidth}{!}{
  \begin{tabular}{|c|c|c|c|c|c|c|}
    \hline
    \multirow{2}{*}{Methods} & \multicolumn{2}{c}{Reversal} \vline&  \multicolumn{2}{c}{Chain} \vline&\multicolumn{2}{c}{Common Cause} \vline\\
    \cline{2-7}
    &Pos&Neg&Pos&Neg&Pos&Neg\\
    \hline
     GPT-3.5&45.9 & 41.2& 43.4&44.7 & 41.6&36.4\\
     GPT-4&49.3 &46.2 &48.9 &43.7& 47.7&48.1\\
    \hline
    RoBERTa&53.9 & 52&56.4 &53.1 & 59.3&56.7\\
    RoBERTa$^{+}$&56.7 &54.9& 58.7& 56.1&52.6&54.6\\
    \hline
    RANK-CP&61.7 & 62.5& 63.4& 61.5& 65.9&63.7\\
     ECPE-2D&63.9 &62.8 &64.6 &61.3 & 63.3&61.9\\
     DialogXL&64.8&60.8&61.9&63.8&65&66.1\\
     EGAT& 68.7& 64.3&68.8 & 64.8&66.2&64.3 \\
     RGAT&69.4&61.7&68.9&66.4&68.3&68.5\\
     DECN&66.7&62.4&70.5&64.3&69.2&66.1\\
     DAG-ERC&71.5&68.2&72.4&64.2&69.3&66.1\\
     \hline
     Ours&76.2&46.1&73.8&41.9&77.2&48.6\\
     \hline 

  \end{tabular}}
  \caption{Results of causal discriminability. 
  The calculation results $=$ count of predicted results 
  that matched the samples $/$ total number of samples $*100$. 
  The three causal models are: Reversal Model: 
  Positive samples (i, j) and negative samples (j, i); 
  Chain Model: Positive samples (i, k) and (k, j), 
  and negative samples (i, j); Common Cause Model: 
  Positive samples (k, i) and (k, j), and negative samples (i, j) 
  or (j, i). The metric ``Pos'' represents the percentage 
  of positive samples, indicating the extraction capability. 
  Higher Pos samples imply a better extraction 
  capability. The metric ``Neg'' represents the percentage 
  of negative samples. A smaller difference between 
  Pos and Neg indicates a weaker causal discriminability.}
  \label{tabcaudis}
\end{table}
We are also concerned about the causal discriminability for 
similar utterances. Table~\ref{tabcaudis} demonstrates that 
in all three different causal models, none of the methods 
were able to distinguish between negative and positive samples. 
Because both negative and positive samples can be fit 
within these three causal models, 
solely from an embedding similarity perspective. 
However, our method significantly decreases the percentage of 
negative samples indicating the effectiveness of 
incorporating implicit cause noise to enhance causal 
discriminative ability.

Additionally, we show the adjacent matrices of our model 
and current SOTA methods in Appendix~\ref{vcg}. 
which indicates that our model 
can more freely explore the relationship between 
different utterances via adjacent matrices shifting 
rather than being limited to a fixed structure 
(e.g., attention module). 

\subsection{Implicit Causes Study (RQ3)} 

\begin{figure}
  \subfigure[without noise]{
    \includegraphics[width=0.48\linewidth]{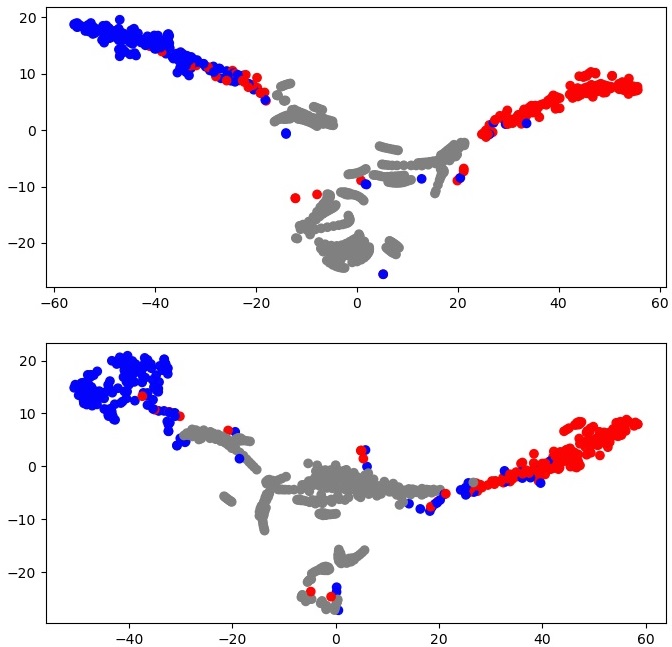}}
  \subfigure[with noise]{
    \includegraphics[width=0.48\linewidth]{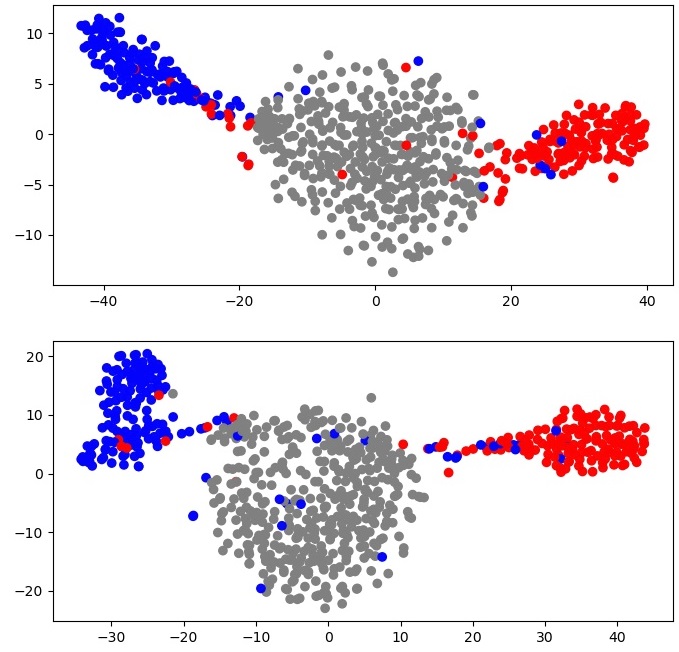}}
  \caption{Visualization of $E$ (upper subfigures) and 
  implicit causes (lower subfigures) with colors in the 
  simulated datasets. The gray cluster means padding utterances 
  in each dialogue, the blue cluster corresponds to the non-emotion utterances, 
  and the red cluster corresponds to emotion utterances. 
   }
  \label{figlatent}
\end{figure}

The latent variable $E$ is intended to represent 
the mentioned implicit causes. Therefore, the global distribution 
of the latent variable $E$ should be approximately equal to the one 
of implicit causes. Although human evaluation labels are better 
for proving reasonable performance, it is intractable to annotate  
implicit causes due to their unobservability. 
We thus trained our model in a synthetic dataset given 
a set of fixed \textit{i.i.d.} implicit causes to observe how the 
$E$ is similar to the ground truth implicit causes distributions.
Figure~\ref{figlatent} (a-b) shows the projection of $E$ and implicit causes, 
respectively, using t-SNE~\citep{knyazev2019understanding}. 
We observe that $E$ and implicit causes 
are both similarly clustered into three parts through 
the distribution properties. 
$E$ is consistent with the implicit causes in the samples 
with or without noise indicating that $E$ successfully 
learns the implicit causes. 

Moreover, in Appendix~\ref{pecicu}, 
we first prove the approximate emotion consistency 
between utterance $U_{t}$ and its implicit causes 
when $U_{t}$ and $U_{i}$ in the emotion-cause pair $(U_{t}$, $U_{i})$ 
do not belong to the same emotion category. 
Then, we demonstrate through the ERC task that by replacing $\hat{H}$ 
with $E$, the emotion consistency provided by implicit causes 
is preserved.

\subsection{Limitations}\label{Results} 

In our model, our method can distinguish between 
$U_{i} \rightarrow U_{j}$ and 
$U_{i} \leftarrow U_{k} \rightarrow U_{j}$. 
However, our method is unable to distinguish between 
$U_{i} \rightarrow U_{j}$ and $U_{i} \leftarrow L \rightarrow U_{j}$, 
where L represents a unobserved variable, called common causes or 
confounders. In Tables~\ref{tabecpeecsr},~\ref{taberc}, and~\ref{tablatenterc}, 
skeletons $\textbf{II}$, $\textbf{III}$, and $\textbf{IV}$ generally 
lag far behind $\textbf{V}$ and $\textbf{VI}$. This unsatisfactory performance of 
these skeletons indicates that excessive adding-edge leads to serious 
confounders. 

Therefore, we proposed a theoretical design for testifying the existing 
of latent confounders: 

\textbf{Confounding between Non-adjacent Nodes}: 
Consider two utterances $U_{i}$ and $U_{j}$ being non-adjacent utterances. 
Let $Pa$ be the union of the parents of $U_{i}$ and $U_{j}$: 
$Pa=U_{i} \cup U_{j}$. If we perform an intervention on $Pa$ 
(i.e., $do(Pa=pa)$), we thus have $U_{i} \perp \!\!\! \perp U_{j}$ if and only 
if there is a latent confounder $L$ such that 
$U_{i} \leftarrow L \rightarrow U_{j}$. 

\textbf{Confounding between Adjacent Nodes}: 
Consider two utterances $U_{i}$ and $U_{j}$ being adjacent utterances: 
$U_{i} \rightarrow U_{j}$. If there are no latent confounders, we 
have $P(U_{j}|U_{i})=P(U_{j}|do(U_{i}=u_{i}))$. 

Indeed, implementing intervention operations on conversation data 
poses a significant challenge. 
Therefore, in our new work, we have proposed general intervention 
writing: $do(X):=Pa(X)=\emptyset$ where $Pa(X)$ denotes the parent set. 
Moreover, the most significant obstacle to further research 
is the lack of a high-quality dataset with complete 
causal relationship labels. Hence, we have constructed a 
simulated dialogue dataset via GPT-4 and plan to make it open soon.

\section{Conclusion}
The results of testing prevalent approaches on the ARC task 
have demonstrated that almost all approaches are unable to 
determine the specific causal relationship that leads to 
the association of two well-fitted embeddings. 
In order to enhance the causal discrimination of existing methods, 
we constructed a SCM with \textit{i.i.d.} noise terms, 
and analyzed the independent conditions that can identify the 
causal relationships between two fitted utterances.  
Moreover, we proposed the \textit{cogn} framework to address the 
unstructured nature of conversation data, 
designed an autoencoder implementation to make implicit cause 
learnable, and created a synthetic dataset with noise labels for 
comprehensive experimental evaluation. While our method still 
has some limitations, such as confounders and the 
inability to scale to all methods, we hope that our theory, 
design, and model can provide valuable insights 
for the broader exploration of this problem 
to demonstrate that our work is \textit{de facto} need 
for identifying causal relationships. 

\bibliography{anthology,custom}

\begin{thebibliography}{7}
\expandafter\ifx\csname natexlab\endcsname\relax\def\natexlab#1{#1}\fi

\bibitem[{Ando and Zhang(2005)}]{Ando2005}
Rie~Kubota Ando and Tong Zhang. 2005.
\newblock A framework for learning predictive structures from multiple tasks
  and unlabeled data.
\newblock \emph{Journal of Machine Learning Research}, 6:1817--1853.

\bibitem[{Andrew and Gao(2007)}]{andrew2007scalable}
Galen Andrew and Jianfeng Gao. 2007.
\newblock Scalable training of {L1}-regularized log-linear models.
\newblock In \emph{Proceedings of the 24th International Conference on Machine
  Learning}, pages 33--40.

\bibitem[{B{\"o}rschinger and
  Johnson(2011)}]{borschinger-johnson-2011-particle}
Benjamin B{\"o}rschinger and Mark Johnson. 2011.
\newblock \href {https://aclanthology.org/U11-1004} {A particle filter
  algorithm for {B}ayesian wordsegmentation}.
\newblock In \emph{Proceedings of the Australasian Language Technology
  Association Workshop 2011}, pages 10--18, Canberra, Australia.

\bibitem[{Goodman et~al.(2016)Goodman, Vlachos, and
  Naradowsky}]{goodman-etal-2016-noise}
James Goodman, Andreas Vlachos, and Jason Naradowsky. 2016.
\newblock \href {https://doi.org/10.18653/v1/P16-1001} {Noise reduction and
  targeted exploration in imitation learning for {A}bstract {M}eaning
  {R}epresentation parsing}.
\newblock In \emph{Proceedings of the 54th Annual Meeting of the Association
  for Computational Linguistics (Volume 1: Long Papers)}, pages 1--11, Berlin,
  Germany. Association for Computational Linguistics.

\bibitem[{Gusfield(1997)}]{Gusfield:97}
Dan Gusfield. 1997.
\newblock \emph{Algorithms on Strings, Trees and Sequences}.
\newblock Cambridge University Press, Cambridge, UK.

\bibitem[{Harper(2014)}]{harper-2014-learning}
Mary Harper. 2014.
\newblock \href {https://aclanthology.org/C14-1001} {Learning from 26
  languages: Program management and science in the babel program}.
\newblock In \emph{Proceedings of {COLING} 2014, the 25th International
  Conference on Computational Linguistics: Technical Papers}, page~1, Dublin,
  Ireland. Dublin City University and Association for Computational
  Linguistics.

\bibitem[{Rasooli and Tetreault(2015)}]{rasooli-tetrault-2015}
Mohammad~Sadegh Rasooli and Joel~R. Tetreault. 2015.
\newblock \href {http://arxiv.org/abs/1503.06733} {Yara parser: {A} fast and
  accurate dependency parser}.
\newblock \emph{Computing Research Repository}, arXiv:1503.06733.
\newblock Version 2.

\end{thebibliography}


\begin{thebibliography}{47}
\expandafter\ifx\csname natexlab\endcsname\relax\def\natexlab#1{#1}\fi

\bibitem[{Agrawal et~al.(2021)Agrawal, Squires, Prasad, and
  Uhler}]{agrawal2021decamfounder}
Raj Agrawal, Chandler Squires, Neha Prasad, and Caroline Uhler. 2021.
\newblock The decamfounder: Non-linear causal discovery in the presence of
  hidden variables.
\newblock \emph{arXiv preprint arXiv:2102.07921}.

\bibitem[{Arora et~al.(2023)Arora, Futami, Tsunoo, Yan, and
  Watanabe}]{arora2023joint}
Siddhant Arora, Hayato Futami, Emiru Tsunoo, Brian Yan, and Shinji Watanabe.
  2023.
\newblock Joint modelling of spoken language understanding tasks with
  integrated dialog history.
\newblock In \emph{ICASSP 2023-2023 IEEE International Conference on Acoustics,
  Speech and Signal Processing (ICASSP)}, pages 1--5. IEEE.

\bibitem[{Bao et~al.(2022)Bao, Ma, Wei, Zhou, and Hu}]{bao2022multi}
Yinan Bao, Qianwen Ma, Lingwei Wei, Wei Zhou, and Songlin Hu. 2022.
\newblock Multi-granularity semantic aware graph model for reducing position
  bias in emotion-cause pair extraction.
\newblock \emph{arXiv preprint arXiv:2205.02132}.

\bibitem[{Busso et~al.(2008)Busso, Bulut, Lee, Kazemzadeh, Mower, Kim, Chang,
  Lee, and Narayanan}]{busso2008iemocap}
Carlos Busso, Murtaza Bulut, Chi-Chun Lee, Abe Kazemzadeh, Emily Mower, Samuel
  Kim, Jeannette~N Chang, Sungbok Lee, and Shrikanth~S Narayanan. 2008.
\newblock Iemocap: Interactive emotional dyadic motion capture database.
\newblock \emph{Language resources and evaluation}, 42(4):335--359.

\bibitem[{Chen et~al.(2023)Chen, Yang, and Li}]{chen2022learning}
Hang Chen, Xinyu Yang, and Chenguang Li. 2023.
\newblock \href {http://arxiv.org/abs/2208.13549} {Learning a general
  clause-to-clause relationships for enhancing emotion-cause pair extraction}.

\bibitem[{Chen et~al.(2018)Chen, Hsu, Kuo, Ku et~al.}]{chen2018emotionlines}
Sheng-Yeh Chen, Chao-Chun Hsu, Chuan-Chun Kuo, Lun-Wei Ku, et~al. 2018.
\newblock Emotionlines: An emotion corpus of multi-party conversations.
\newblock \emph{arXiv preprint arXiv:1802.08379}.

\bibitem[{Cheng et~al.(2022)Cheng, Guo, Moraffah, Sheth, Candan, and
  Liu}]{cheng2022evaluation}
Lu~Cheng, Ruocheng Guo, Raha Moraffah, Paras Sheth, Kasim~Selcuk Candan, and
  Huan Liu. 2022.
\newblock Evaluation methods and measures for causal learning algorithms.
\newblock \emph{IEEE Transactions on Artificial Intelligence}.

\bibitem[{Colombo et~al.(2012)Colombo, Maathuis, Kalisch, and
  Richardson}]{colombo2012learning}
Diego Colombo, Marloes~H Maathuis, Markus Kalisch, and Thomas~S Richardson.
  2012.
\newblock Learning high-dimensional directed acyclic graphs with latent and
  selection variables.
\newblock \emph{The Annals of Statistics}, pages 294--321.

\bibitem[{Ding et~al.(2020)Ding, Xia, and Yu}]{ding-etal-2020-ecpe}
Zixiang Ding, Rui Xia, and Jianfei Yu. 2020.
\newblock \href {https://doi.org/10.18653/v1/2020.acl-main.288} {{ECPE}-2{D}:
  Emotion-cause pair extraction based on joint two-dimensional representation,
  interaction and prediction}.
\newblock In \emph{Proceedings of the 58th Annual Meeting of the Association
  for Computational Linguistics}, pages 3161--3170, Online. Association for
  Computational Linguistics.

\bibitem[{Feng et~al.(2022)Feng, Lubis, Geishauser, Lin, Heck, van Niekerk, and
  Ga{\v{s}}ic}]{fengemowoz}
Shutong Feng, Nurul Lubis, Christian Geishauser, Hsien-chin Lin, Michael Heck,
  Carel van Niekerk, and Milica Ga{\v{s}}ic. 2022.
\newblock Emowoz: A large-scale corpus and labelling scheme for emotion
  recognition in task-oriented dialogue systems.

\bibitem[{Ghosal et~al.(2019)Ghosal, Majumder, Poria, Chhaya, and
  Gelbukh}]{ghosal-etal-2019-dialoguegcn}
Deepanway Ghosal, Navonil Majumder, Soujanya Poria, Niyati Chhaya, and
  Alexander Gelbukh. 2019.
\newblock \href {https://doi.org/10.18653/v1/D19-1015} {{D}ialogue{GCN}: A
  graph convolutional neural network for emotion recognition in conversation}.
\newblock In \emph{Proceedings of the 2019 Conference on Empirical Methods in
  Natural Language Processing and the 9th International Joint Conference on
  Natural Language Processing (EMNLP-IJCNLP)}, pages 154--164, Hong Kong,
  China. Association for Computational Linguistics.

\bibitem[{Ishiwatari et~al.(2021)Ishiwatari, Yasuda, Miyazaki, and
  Goto}]{ishiwatari-etal-2020-relation}
Taichi Ishiwatari, Yuki Yasuda, Taro Miyazaki, and Jun Goto. 2021.
\newblock \href {https://doi.org/10.18653/v1/2020.emnlp-main.597}
  {Relation-aware graph attention networks with relational position encodings
  for emotion recognition in conversations}.
\newblock In \emph{Proceedings of the 2021 Conference on Empirical Methods in
  Natural Language Processing (EMNLP)}, pages 7360--7370, Online. Association
  for Computational Linguistics.

\bibitem[{Joyce(2011)}]{joyce2011kullback}
James~M Joyce. 2011.
\newblock Kullback-leibler divergence.
\newblock In \emph{International encyclopedia of statistical science}, pages
  720--722. Springer.

\bibitem[{Kasneci et~al.(2023)Kasneci, Se{\ss}ler, K{\"u}chemann, Bannert,
  Dementieva, Fischer, Gasser, Groh, G{\"u}nnemann, H{\"u}llermeier
  et~al.}]{kasneci2023chatgpt}
Enkelejda Kasneci, Kathrin Se{\ss}ler, Stefan K{\"u}chemann, Maria Bannert,
  Daryna Dementieva, Frank Fischer, Urs Gasser, Georg Groh, Stephan
  G{\"u}nnemann, Eyke H{\"u}llermeier, et~al. 2023.
\newblock Chatgpt for good? on opportunities and challenges of large language
  models for education.
\newblock \emph{Learning and Individual Differences}, 103:102274.

\bibitem[{Kingma and Welling(2014)}]{2014Auto}
D.~P. Kingma and M.~Welling. 2014.
\newblock Auto-encoding variational bayes.
\newblock \emph{arXiv.org}.

\bibitem[{Knyazev et~al.(2019)Knyazev, Taylor, and
  Amer}]{knyazev2019understanding}
Boris Knyazev, Graham~W Taylor, and Mohamed Amer. 2019.
\newblock Understanding attention and generalization in graph neural networks.
\newblock \emph{Advances in neural information processing systems}, 32.

\bibitem[{Krych-Appelbaum et~al.(2007)Krych-Appelbaum, Law, Jones, Barnacz,
  Johnson, and Keenan}]{krych2007think}
Meredyth Krych-Appelbaum, Julie~Banzon Law, Dayna Jones, Allyson Barnacz,
  Amanda Johnson, and Julian~Paul Keenan. 2007.
\newblock “i think i know what you mean”: The role of theory of mind in
  collaborative communication.
\newblock \emph{Interaction Studies}, 8(2):267--280.

\bibitem[{Li et~al.(2017)Li, Su, Shen, Li, Cao, and
  Niu}]{li-etal-2017-dailydialog}
Yanran Li, Hui Su, Xiaoyu Shen, Wenjie Li, Ziqiang Cao, and Shuzi Niu. 2017.
\newblock \href {https://aclanthology.org/I17-1099} {{D}aily{D}ialog: A
  manually labelled multi-turn dialogue dataset}.
\newblock In \emph{Proceedings of the Eighth International Joint Conference on
  Natural Language Processing (Volume 1: Long Papers)}, pages 986--995, Taipei,
  Taiwan. Asian Federation of Natural Language Processing.

\bibitem[{Lian et~al.(2021)Lian, Liu, and Tao}]{lian2021decn}
Zheng Lian, Bin Liu, and Jianhua Tao. 2021.
\newblock Decn: Dialogical emotion correction network for conversational
  emotion recognition.
\newblock \emph{Neurocomputing}, 454:483--495.

\bibitem[{Liu et~al.(2019)Liu, Ott, Goyal, Du, Joshi, Chen, Levy, Lewis,
  Zettlemoyer, and Stoyanov}]{liu2019roberta}
Yinhan Liu, Myle Ott, Naman Goyal, Jingfei Du, Mandar Joshi, Danqi Chen, Omer
  Levy, Mike Lewis, Luke Zettlemoyer, and Veselin Stoyanov. 2019.
\newblock Roberta: A robustly optimized bert pretraining approach.
\newblock \emph{arXiv preprint arXiv:1907.11692}.

\bibitem[{Majumder et~al.(2019)Majumder, Poria, Hazarika, Mihalcea, Gelbukh,
  and Cambria}]{majumder2019dialoguernn}
Navonil Majumder, Soujanya Poria, Devamanyu Hazarika, Rada Mihalcea, Alexander
  Gelbukh, and Erik Cambria. 2019.
\newblock Dialoguernn: An attentive rnn for emotion detection in conversations.
\newblock In \emph{Proceedings of the AAAI conference on artificial
  intelligence}, volume~33, pages 6818--6825.

\bibitem[{Ni(2023)}]{ni2023attention}
Jinjie Ni. 2023.
\newblock Attention mechanism optimization for sub-symbolic-based and
  neural-symbolic-based natural language processing.

\bibitem[{Nogueira et~al.(2022)Nogueira, Pugnana, Ruggieri, Pedreschi, and
  Gama}]{nogueira2022methods}
Ana~Rita Nogueira, Andrea Pugnana, Salvatore Ruggieri, Dino Pedreschi, and
  Jo{\~a}o Gama. 2022.
\newblock Methods and tools for causal discovery and causal inference.
\newblock \emph{Wiley Interdisciplinary Reviews: Data Mining and Knowledge
  Discovery}, 12(2):e1449.

\bibitem[{Ogarrio et~al.(2016)Ogarrio, Spirtes, and Ramsey}]{ogarrio2016hybrid}
Juan~Miguel Ogarrio, Peter Spirtes, and Joe Ramsey. 2016.
\newblock A hybrid causal search algorithm for latent variable models.
\newblock In \emph{Conference on probabilistic graphical models}, pages
  368--379. PMLR.

\bibitem[{Ong et~al.(2015)Ong, Zaki, and Goodman}]{ong2015affective}
Desmond~C Ong, Jamil Zaki, and Noah~D Goodman. 2015.
\newblock Affective cognition: Exploring lay theories of emotion.
\newblock \emph{Cognition}, 143:141--162.

\bibitem[{Ong et~al.(2019)Ong, Zaki, and Goodman}]{ong2019computational}
Desmond~C Ong, Jamil Zaki, and Noah~D Goodman. 2019.
\newblock Computational models of emotion inference in theory of mind: A review
  and roadmap.
\newblock \emph{Topics in cognitive science}, 11(2):338--357.

\bibitem[{Pereira et~al.(2023)Pereira, Moniz, and Carvalho}]{pereira2022deep}
Patr{\'\i}cia Pereira, Helena Moniz, and Joao~Paulo Carvalho. 2023.
\newblock Deep emotion recognition in textual conversations: A survey.
\newblock \emph{arXiv preprint arXiv:2211.09172}.

\bibitem[{Poria et~al.(2019)Poria, Hazarika, Majumder, Naik, Cambria, and
  Mihalcea}]{poria-etal-2019-meld}
Soujanya Poria, Devamanyu Hazarika, Navonil Majumder, Gautam Naik, Erik
  Cambria, and Rada Mihalcea. 2019.
\newblock \href {https://doi.org/10.18653/v1/P19-1050} {{MELD}: A multimodal
  multi-party dataset for emotion recognition in conversations}.
\newblock In \emph{Proceedings of the 57th Annual Meeting of the Association
  for Computational Linguistics}, pages 527--536, Florence, Italy. Association
  for Computational Linguistics.

\bibitem[{Poria et~al.(2021)Poria, Majumder, Hazarika, Ghosal, Bhardwaj, Jian,
  Hong, Ghosh, Roy, Chhaya et~al.}]{poria2021recognizing}
Soujanya Poria, Navonil Majumder, Devamanyu Hazarika, Deepanway Ghosal, Rishabh
  Bhardwaj, Samson Yu~Bai Jian, Pengfei Hong, Romila Ghosh, Abhinaba Roy,
  Niyati Chhaya, et~al. 2021.
\newblock Recognizing emotion cause in conversations.
\newblock \emph{Cognitive Computation}, 13(5):1317--1332.

\bibitem[{Sanchez-Romero et~al.(2019)Sanchez-Romero, Ramsey, Zhang, Glymour,
  Huang, and Glymour}]{sanchez2019estimating}
Ruben Sanchez-Romero, Joseph~D Ramsey, Kun Zhang, Madelyn~RK Glymour, Biwei
  Huang, and Clark Glymour. 2019.
\newblock Estimating feedforward and feedback effective connections from fmri
  time series: Assessments of statistical methods.
\newblock \emph{Network Neuroscience}, 3(2):274--306.

\bibitem[{Shen et~al.(2022)Shen, Chen, Quan, and Xie}]{Shen2021DialogXLAX}
Weizhou Shen, Junqing Chen, Xiaojun Quan, and Zhixiang Xie. 2022.
\newblock Dialogxl: All-in-one xlnet for multi-party conversation emotion
  recognition.
\newblock In \emph{AAAI}.

\bibitem[{Shen et~al.(2021)Shen, Wu, Yang, and Quan}]{shen-etal-2021-directed}
Weizhou Shen, Siyue Wu, Yunyi Yang, and Xiaojun Quan. 2021.
\newblock \href {https://doi.org/10.18653/v1/2021.acl-long.123} {Directed
  acyclic graph network for conversational emotion recognition}.
\newblock In \emph{Proceedings of the 59th Annual Meeting of the Association
  for Computational Linguistics and the 11th International Joint Conference on
  Natural Language Processing (Volume 1: Long Papers)}, pages 1551--1560,
  Online. Association for Computational Linguistics.

\bibitem[{Shimizu and Bollen(2014)}]{shimizu2014bayesian}
Shohei Shimizu and Kenneth Bollen. 2014.
\newblock Bayesian estimation of causal direction in acyclic structural
  equation models with individual-specific confounder variables and
  non-gaussian distributions.
\newblock \emph{J. Mach. Learn. Res.}, 15(1):2629--2652.

\bibitem[{Shimizu et~al.(2006)Shimizu, Hoyer, Hyv{\"a}rinen, Kerminen, and
  Jordan}]{shimizu2006linear}
Shohei Shimizu, Patrik~O Hoyer, Aapo Hyv{\"a}rinen, Antti Kerminen, and Michael
  Jordan. 2006.
\newblock A linear non-gaussian acyclic model for causal discovery.
\newblock \emph{Journal of Machine Learning Research}, 7(10).

\bibitem[{Shirai et~al.(2023)Shirai, Kameko, and Mori}]{shirai2023towards}
Keisuke Shirai, Hirotaka Kameko, and Shinsuke Mori. 2023.
\newblock Towards flow graph prediction of open-domain procedural texts.
\newblock \emph{arXiv preprint arXiv:2305.19497}.

\bibitem[{Sidera et~al.(2018)Sidera, Perpi{\~n}{\`a}, Serrano, and
  Rostan}]{sidera2018theory}
Francesc Sidera, Georgina Perpi{\~n}{\`a}, J{\`e}ssica Serrano, and Carles
  Rostan. 2018.
\newblock Why is theory of mind important for referential communication?
\newblock \emph{Current Psychology}, 37:82--97.

\bibitem[{Spirtes et~al.(2000)Spirtes, Glymour, Scheines, Kauffman, Aimale, and
  Wimberly}]{spirtes2000constructing}
Pater Spirtes, Clark Glymour, Richard Scheines, Stuart Kauffman, Valerio
  Aimale, and Frank Wimberly. 2000.
\newblock Constructing bayesian network models of gene expression networks from
  microarray data.

\bibitem[{Squires et~al.(2022)Squires, Yun, Nichani, Agrawal, and
  Uhler}]{squires2022causal}
Chandler Squires, Annie Yun, Eshaan Nichani, Raj Agrawal, and Caroline Uhler.
  2022.
\newblock Causal structure discovery between clusters of nodes induced by
  latent factors.
\newblock In \emph{Conference on Causal Learning and Reasoning}, pages
  669--687. PMLR.

\bibitem[{Thakur et~al.(2023)Thakur, Shahu, and Gupta}]{thakur2023audio}
Palash Thakur, Ronit Shahu, and Vikas Gupta. 2023.
\newblock Audio and text-based emotion recognition system using deep learning.
\newblock In \emph{Advances in Signal Processing, Embedded Systems and IoT:
  Proceedings of Seventh ICMEET-2022}, pages 447--459. Springer.

\bibitem[{Thost and Chen(2021)}]{thost2021directed}
Veronika Thost and Jie Chen. 2021.
\newblock Directed acyclic graph neural networks.
\newblock \emph{arXiv preprint arXiv:2101.07965}.

\bibitem[{Uymaz and Metin(2022)}]{uymaz2022vector}
Hande~Aka Uymaz and Senem~Kumova Metin. 2022.
\newblock Vector based sentiment and emotion analysis from text: A survey.
\newblock \emph{Engineering Applications of Artificial Intelligence},
  113:104922.

\bibitem[{Veli{\v{c}}kovi{\'c} et~al.(2017)Veli{\v{c}}kovi{\'c}, Cucurull,
  Casanova, Romero, Lio, and Bengio}]{velivckovic2017graph}
Petar Veli{\v{c}}kovi{\'c}, Guillem Cucurull, Arantxa Casanova, Adriana Romero,
  Pietro Lio, and Yoshua Bengio. 2017.
\newblock Graph attention networks.
\newblock \emph{arXiv preprint arXiv:1710.10903}.

\bibitem[{Wei et~al.(2020)Wei, Zhao, and Mao}]{wei-etal-2020-effective}
Penghui Wei, Jiahao Zhao, and Wenji Mao. 2020.
\newblock \href {https://doi.org/10.18653/v1/2020.acl-main.289} {Effective
  inter-clause modeling for end-to-end emotion-cause pair extraction}.
\newblock In \emph{Proceedings of the 58th Annual Meeting of the Association
  for Computational Linguistics}, pages 3171--3181, Online. Association for
  Computational Linguistics.

\bibitem[{Xia and Ding(2019)}]{xia-ding-2019-emotion}
Rui Xia and Zixiang Ding. 2019.
\newblock \href {https://doi.org/10.18653/v1/P19-1096} {Emotion-cause pair
  extraction: A new task to emotion analysis in texts}.
\newblock In \emph{Proceedings of the 57th Annual Meeting of the Association
  for Computational Linguistics}, pages 1003--1012, Florence, Italy.
  Association for Computational Linguistics.

\bibitem[{Ye et~al.(2023)Ye, Wang, and Chen}]{ye2023msmix}
Mao Ye, Haitao Wang, and Zheqian Chen. 2023.
\newblock Msmix: An interpolation-based text data augmentation method manifold
  swap mixup.
\newblock \emph{arXiv preprint arXiv:2305.19617}.

\bibitem[{Zahiri and Choi(2018)}]{zahiri:18a}
Sayyed Zahiri and Jinho~D. Choi. 2018.
\newblock \href {https://sites.google.com/view/affcon18} {{Emotion Detection on
  TV Show Transcripts with Sequence-based Convolutional Neural Networks}}.
\newblock In \emph{Proceedings of the AAAI Workshop on Affective Content
  Analysis}, AFFCON'18, pages 44--51, New Orleans, LA.

\bibitem[{Zhang et~al.(2019)Zhang, Wu, Sun, Li, Zhu, and
  Zhou}]{zhang2019modeling}
Dong Zhang, Liangqing Wu, Changlong Sun, Shoushan Li, Qiaoming Zhu, and Guodong
  Zhou. 2019.
\newblock Modeling both context-and speaker-sensitive dependence for emotion
  detection in multi-speaker conversations.
\newblock In \emph{IJCAI}, pages 5415--5421.

\end{thebibliography}
\bibliographystyle{acl_natbib}

\appendix

\section{Proof of Definition 2}\label{pod2}

Let $X$ and $Y$ be two variables in an SCM, 
with their respective noise terms denoted as $E_{X}$ and $E_{Y}$ 
(where $E_{X}$ and $E_{Y}$ are mutually independent). 
Let $\hat{X}$ and $\hat{Y}$ represent the fitted values of 
$X$ and $Y$ w.r.t. each other:  $\hat{X}=\lambda Y$ and 
$\hat{Y}=\frac{1}{\lambda} X$. The residual terms 
between the fitted values and the true values are denoted as 
$\Sigma_{X}=X-\hat{X}$ and $\Sigma_{Y}=Y-\hat{Y}$. 
The true strength of $Y \rightarrow X$ is $k$. 

Hence, if the SCM only contains two variables writing:
\begin{equation}
  Y=E_{Y}
  \label{eqnp1}
 \end{equation}
 \begin{equation}
  X=kY+E_{X}
  \label{eqnp2}
 \end{equation}
The residual terms could write: 
\begin{equation}
  \Sigma_{X}=X-\lambda (\frac{1}{k}(X-E_{X}))
  \label{eqnp3}
 \end{equation}
 \begin{equation}
  \Sigma_{Y}=Y-\frac{1}{\lambda}(ky+E_{X})
  \label{eqnp4}
 \end{equation}
Then, if the true causal relationship is from $Y$ to $X$, 
$\lambda=k$. 
$\Sigma_{X}$ does not contain the term of $E_{Y}$ while 
$\Sigma_{Y}$ contains the term of $E_{X}$. 
We could obtain the independence of residual terms writting: 
\begin{equation}
  \Sigma_{X}=\lambda E_{X} \perp \!\!\! \perp Y
  \label{eqnp5}
 \end{equation}
 \begin{equation}
  \Sigma_{Y}=\frac{1}{\lambda} E_{X} \not \! \perp \!\!\! \perp X
  \label{eqnp6}
 \end{equation}
and vice versa. Therefore, we could obtain the independence condition: 
\begin{itemize}
  \item $\Sigma_{X} \perp \!\!\! \perp Y, 
  \Sigma_{Y} \not \! \perp \!\!\! \perp X
  \Rightarrow Y \rightarrow X$

  \item $\Sigma_{X} \not \! \perp \!\!\! \perp Y, 
  \Sigma_{Y} \perp \!\!\! \perp X
  \Rightarrow X \rightarrow Y$
\end{itemize}
Furthermore, there may exist a set of independence: 
$\Sigma_{X} \not \! \perp \!\!\! \perp Y$, 
$\Sigma_{Y} \not \! \perp \!\!\! \perp X$. We would like to assume 
that there is a latent variable $L$, for this situation, 
constructing two relationships 
$L \rightarrow X$ and $L \rightarrow Y$. Then we obtain: 
$\Sigma_{L} \not \! \perp \!\!\! \perp X$, 
$\Sigma_{L} \not \! \perp \!\!\! \perp Y$. 
By utilizing the transitivity of conditional independence, 
we can establish $X \not \! \perp \!\!\! \perp Y$, and finally 
acheive the situation 
$\Sigma_{X} \not \! \perp \!\!\! \perp Y$, 
$\Sigma_{Y} \not \! \perp \!\!\! \perp X$. We likewise assume a 
latent variable $L$ establishing 
$X \rightarrow L$ and $Y \rightarrow L$ 
for the opposite situation where 
$\Sigma_{X} \perp \!\!\! \perp Y$, $\Sigma_{Y} \perp \!\!\! \perp X$, 
and $X$, $Y$ are two isolated variables 
in SCM. From the above independence 
conditions, we could obtain: 
$\Sigma_{L}  \perp \!\!\! \perp X$, 
$\Sigma_{L}  \perp \!\!\! \perp Y$. Due to the graph structure of 
SCM, we could obtain: $\Sigma_{X} \perp \!\!\! \perp Y, 
\Sigma_{Y} \perp \!\!\! \perp X \Rightarrow X \perp \!\!\! \perp Y$. 
Considering the residual terms, we finally obtain: 
$X \not \! \perp \!\!\! \perp \Sigma_{X}$ and $X \perp \!\!\! \perp Y 
\Rightarrow \Sigma_{X} \perp \!\!\! \perp Y$ and 
$Y \not \! \perp \!\!\! \perp \Sigma_{Y}$ and $Y \perp \!\!\! \perp X 
\Rightarrow \Sigma_{Y} \perp \!\!\! \perp X$. 

Hence, we could obtain additional two independence conditions: 

\begin{itemize}

  \item $\Sigma_{X} \not \! \perp \!\!\! \perp Y, 
  \Sigma_{Y} \not \! \perp \!\!\! \perp X
  \Rightarrow L \rightarrow X, L \rightarrow Y$

  \item $\Sigma_{X} \perp \!\!\! \perp Y, 
  \Sigma_{Y} \perp \!\!\! \perp X
  \Rightarrow X \rightarrow L, Y \rightarrow L$

\end{itemize}

Based on the independence conditions of 2-variables SCM, we could 
extend it to the general SCM including more than 2 variables. 
Given any two variables in a SCM, we could testify to the independence 
condition and finally orientate via the whole SCM.

\section{Hypotheses and Algorithms for Skeletons}\label{has}

\textbf{Hypothesis 0.} \textit{$\forall U_{i} \in \mathcal{D}$, it 
has the same causal skeleton as other utterances.}

By regarding \textbf{Hypothesis 0} as the prior knowledge, 
a common causal skeleton containing a target variable and 
a fixed number of related variables can reason about 
the relations between the target utterance and other considered utterances. 
We denote this skeleton of $U_{t}$ by $S(U_{t})$. There are 
$\forall U_{i}, U_{j}\in \mathcal{D}, S(U_{i}) = S(U_{j})$. 
 
Additionally, there are some other empirical hypotheses from the above 
approaches. These hypotheses can be divided into two categories: 
one is about the ``order'' of utterances (\textbf{Hypotheses 1, 2, 3}), 
and the other is about intermingling dynamics among the interlocutors 
(\textbf{Hypotheses 4, 5, 6}).

\textbf{Hypothesis 1.}~\citep{majumder2019dialoguernn} 
\textit{Under the sequential order, the target 
utterance receives information only from the previous utterance.}

\textbf{Hypothesis 2.}~\citep{wei-etal-2020-effective}  
\textit{Under the graph order, 
the target utterance receives information from all other utterances.}

\textbf{Hypothesis 3.}~\citep{ghosal-etal-2019-dialoguegcn}
\textit{Under the local graph order, 
target utterance receives local information from k surround utterances.}

\textbf{Hypothesis 4.}~\citep{zhang2019modeling}
\textit{The influence between two utterances can be discriminated 
by whether the two utterances belong to the same speaker identity. }

\textbf{Hypothesis 5.}~\citep{lian2021decn} 
\textit{Target utterance only receives information from  
the predecessor utterances.}

\textbf{Hypothesis 6.}~\citep{shen-etal-2021-directed}
\textit{Between two utterances both related to the target utterance, 
there is also information passing, often dubbed as a partial order.}

\begin{table}
  \footnotesize
  \centering
  \begin{tabular}{|c|c|c|}
    \hline
    \textbf{Category}&\textbf{Hypothesis} & \textbf{Original work}\\
    \hline
    $\textbf{I}$ & 1 &~\citet{majumder2019dialoguernn}\\
    $\textbf{II}$ & 2 &~\citet{velivckovic2017graph}\\
    $\textbf{III}$ & 2,4 &\citet{chen2022learning}\\
    $\textbf{IV}$ & 3, 4 &~\citet{ghosal-etal-2019-dialoguegcn}\\
    $\textbf{V}$ & 3($k=1$), 4, 5 &~\citet{lian2021decn}\\
    $\textbf{VI}$ & 3, 4, 5, 6 &~\citet{shen-etal-2021-directed}\\
    \hline
  \end{tabular}
  \caption{ Statistics of 6 \textit{cogn} skeletons. We detailed 
  the hypotheses each \textit{cogn} skeleton adopted 
  and the original works from which we designed them.}
  \label{tabskeleton}
\end{table}

A \textit{cogn} skeleton is denoted by 
$\mathcal{H} =(\mathcal{V}, \mathcal{E}, \mathcal{M})$.
The $\mathcal{V}={U_{1},U_{2},U_{3},...,U_{N}}$ represents 
a set of utterances in a conversation, 
and the edge $(i, j, m_{i,j})\in \mathcal{E}$ denotes the 
influence from $U_{i}$ to $U_{j}$, where $m_{i,j}\in \mathcal{M}$ 
is the type of the edge depending on whether $U_{i}$ and $U_{j}$ 
belong to one and the same speaker. Thus $\mathcal{M}={0,1}$, where 
$1$ for that they are the same speaker and $0$ for different. 
Then we denote the speaker type of $U_{i}$ by a function $p(U_{i})$. 
At last, 
we show the process of building 6 \textit{cogn} skeletons in Algorithms $1-6$.

\begin{algorithm}
  \KwInput{$\mathcal{D}$, $p(\cdot)$, $k$}
  \KwOutput{$\mathcal{H} =(\mathcal{V}, \mathcal{E})$}
  $\mathcal{V} \leftarrow {U_{1},U_{2},U_{3},...,U_{N}}$\\
  $\mathcal{E} \leftarrow \emptyset $\\
  \ForAll{$i \in {2, 3, \ldots, N-1}$}
  {
        $\mathcal{E} \leftarrow \mathcal{E}\cup {(i, i+1)}$\\
  } 
  return $\mathcal{H} =(\mathcal{V}, \mathcal{E})$
\caption{Buliding $\textbf{I}$ \textit{cogn} skeleton}
\end{algorithm}

\begin{algorithm}
  \KwInput{$\mathcal{D}$, $p(\cdot)$, $k$}
  \KwOutput{$\mathcal{H} =(\mathcal{V}, \mathcal{E})$}
  $\mathcal{V} \leftarrow {U_{1},U_{2},U_{3},...,U_{N}}$\\
  $\mathcal{E} \leftarrow \emptyset $\\
  \ForAll{$i \in {2, 3, \ldots, N}$}
  {     
        \ForAll{$j \in {2,3,\ldots,N}$}
        {
          \If{$i != j$}
          {
            $\mathcal{E} \leftarrow \mathcal{E}\cup {(j, i)}$\\
          }
          \Else
          {
            $Continue$
          }
        }
        
  } 
  return $\mathcal{H} =(\mathcal{V}, \mathcal{E})$
\caption{Buliding $\textbf{II}$ \textit{cogn} skeleton}
\end{algorithm}

\begin{algorithm}
  \KwInput{$\mathcal{D}$, $p(\cdot)$, $k$}
  \KwOutput{$\mathcal{H} =(\mathcal{V}, \mathcal{E}, \mathcal{M})$}
  $\mathcal{V} \leftarrow {U_{1},U_{2},U_{3},...,U_{N}}$\\
  $\mathcal{E} \leftarrow \emptyset $\\
  $\mathcal{M} \leftarrow {0, 1}$\\
  \ForAll{$i \in {2, 3, \ldots, N}$}
  {
    \ForAll{$j \in {2,3,\ldots,N}$}
    {
      \If{$p(U_{j})=p(U_{i}) $ and $i != j$}
      {
        $\mathcal{E} \leftarrow \mathcal{E}\cup {(j, i, 1)}$\\
      }
      \ElseIf{$p(U_{j})!=p(U_{i}) $ and $i != j$}
      {
        $\mathcal{E} \leftarrow \mathcal{E}\cup {(j, i, 0)}$
      }
      \Else
      {
        $Continue$
      }
    }
    
  } 
  return $\mathcal{H} =(\mathcal{V}, \mathcal{E}, \mathcal{M})$
\caption{Buliding $\textbf{III}$ \textit{cogn} skeleton}
\end{algorithm}

\begin{algorithm}
  \KwInput{$\mathcal{D}$, $p(\cdot)$, $k$}
  \KwOutput{$\mathcal{H} =(\mathcal{V}, \mathcal{E}, \mathcal{M})$}
  $\mathcal{V} \leftarrow {U_{1},U_{2},U_{3},...,U_{N}}$\\
  $\mathcal{E} \leftarrow \emptyset $\\
  $\mathcal{M} \leftarrow {0, 1}$\\
  \ForAll{$i \in {2, 3, \ldots, N}$}
  {
    \ForAll{$j \in {2,3,\ldots,N}$}
    {
      \If{$p(U_{j})=p(U_{i}) $ and $0<|i-j|<k$}
      {
        $\mathcal{E} \leftarrow \mathcal{E}\cup {(j, i, 1)}$\\
      }
      \ElseIf{$p(U_{j})!=p(U_{i}) $ and $0<|i-j|<k$}
      {
        $\mathcal{E} \leftarrow \mathcal{E}\cup {(j, i, 0)}$
      }
      \Else
      {
        $Continue$
      }
    }
    
  } 
  return $\mathcal{H} =(\mathcal{V}, \mathcal{E}, \mathcal{M})$
\caption{Buliding $\textbf{IV}$ \textit{cogn} skeleton}
\end{algorithm}

\begin{algorithm}
  \KwInput{$\mathcal{D}$, $p(\cdot)$, $k$}
  \KwOutput{$\mathcal{H} =(\mathcal{V}, \mathcal{E}, \mathcal{M})$}
  $\mathcal{V} \leftarrow {U_{1},U_{2},U_{3},...,U_{N}}$\\
  $\mathcal{E} \leftarrow \emptyset $\\
  $\mathcal{M} \leftarrow {0, 1}$\\
  \ForAll{$i \in {2, 3, \ldots, N}$}
  {

    $\gamma \leftarrow i-1$\\
    
      \If{$p(U_{\gamma})=p(U_{i})$}
      {
        $\mathcal{E} \leftarrow \mathcal{E}\cup {(\gamma, i, 1)}$\\
        
      }
      \Else
      {
        $\mathcal{E} \leftarrow \mathcal{E}\cup {(\gamma, i, 0)}$
      }
      $\gamma \leftarrow \gamma-1$
    
  } 
  return $\mathcal{H} =(\mathcal{V}, \mathcal{E}, \mathcal{M})$
\caption{Buliding $\textbf{V}$ \textit{cogn} skeleton}
\end{algorithm}

\begin{figure*}
  \includegraphics[width=0.95\linewidth]{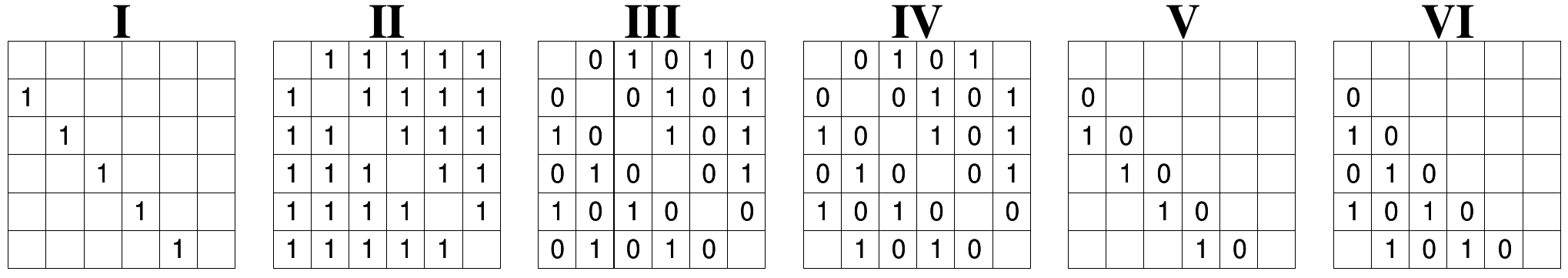}
  \caption{Adjacency matrices towards 6 \textit{cogn} skeletons when $k=2$. 
  $(i,j)\neq None$ represents that $U_{i}$ is influenced by $U_{j}$. }
  \label{figskeleton_case}
\end{figure*}

\begin{algorithm}
  \KwInput{$\mathcal{D}$, $p(\cdot)$, $k$}
  \KwOutput{$\mathcal{H} =(\mathcal{V}, \mathcal{E}, \mathcal{M})$}
  $\mathcal{V} \leftarrow {U_{1},U_{2},U_{3},...,U_{N}}$\\
  $\mathcal{E} \leftarrow \emptyset $\\
  $\mathcal{M} \leftarrow {0, 1}$\\
  \ForAll{$i \in {2, 3, \ldots, N}$}
  {
    $c \leftarrow 0$\\
    $\gamma \leftarrow i-1$\\
    \While{$\gamma>0$ and $c<k$}
    {
      \If{$p(U_{\gamma})=p(U_{i})$}
      {
        $\mathcal{E} \leftarrow \mathcal{E}\cup {(\gamma, i, 1)}$\\
        $c \leftarrow c+1$
      }
      \Else
      {
        $\mathcal{E} \leftarrow \mathcal{E}\cup {(\gamma, i, 0)}$
      }
      $\gamma \leftarrow \gamma-1$
    }
  } 
  return $\mathcal{H} =(\mathcal{V}, \mathcal{E}, \mathcal{M})$
\caption{Buliding $\textbf{VI}$ \textit{cogn} skeleton}
\end{algorithm}

Finally, in Figure~\ref{figskeleton_case}, we show the adjacency 
matrix of each \textit{cogn} skeleton by 
inputting a binary alternating conversation case with 6 utterances. But 
note that adjacency can not indicate all the differences among these 
skeletons, for example, Hypothesis 6 takes effect when 
the model learns the relationship based on the $\textbf{VI}$ skeleton.

\section{Datasets}\label{D}

\textbf{DailyDialog}~\citep{li-etal-2017-dailydialog}:
A Human-written dialogs dataset with 7 emotion labels (\textit{neutral}, 
\textit{happiness}, \textit{surprise}, \textit{sadness}, 
\textit{anger}, \textit{disgust}, and \textit{fear}). We follow 
~\citet{shen-etal-2021-directed} to regard utterance turns as speaker turns. 

\textbf{MELD}~\citep{poria-etal-2019-meld}: 
A multimodel ERC dataset with 7 emotion labels as the same as DailyDialog. 

\textbf{EmoryNLP}~\citep{zahiri:18a}:
A TV show scripts dataset with 7 emotion labels (\textit{neutral}, 
\textit{sad}, \textit{mad}, \textit{scared}, \textit{powerful}, 
\textit{peaceful}, \textit{joyful}). 

\textbf{IEMOCAP}~\citep{busso2008iemocap}: 
A multimodel ERC dataset with 9 emotion labels (\textit{neutral}, 
\textit{happy}, \textit{sad}, \textit{angry}, \textit{frustrated}, 
\textit{excited}, \textit{surprised}, \textit{disappointed}, and \textit{fear}). 
However, models in ERC field are often evaluated on samples with 
first six emotions due to the too few samples of latter three emotions. 
20 dialogues for validation set is following~\citet{shen-etal-2021-directed}. 

\textbf{RECCON}~\citep{poria2021recognizing}: 
The first dataset for emotion cause recognition of conversation 
including RECCON-DD and RECCON-IE (a subset emulating an out-of-distribution 
generalization test). RECCON-DD includes 5380 labeled ECPs and 5 cause 
spans (\textit{no-context}, \textit{inter-personal}, \textit{self-contagion}, 
\textit{hybrid}, and \textit{latent}). 

\textbf{Synthetic dataset}: 
We create a synthetic dataset by following the benchmark of the 
causal discovery field~\cite{agrawal2021decamfounder,squires2022causal}. 
To minimize sample bias, we did not randomly draw causal graphs as 
samples. Inversely, the number of samples in the synthetic dataset 
and the number of utterances and labels per sample are restricted to be consistent with RECCON.
We use Causal Additive Models (CAMs), Specifically SCM structure for our datasets. 
As shown in Algorithm~\ref{aldata}, first, we assume that each $i.i.d.$ implicit causes
$E \sim \|^{50}  \mathcal{N} (1,1)$ if it is an emotion utterance in the original dataset, 
and $E \sim \|^{50} \mathcal{N} (-1,1)$ if it is not. Then, we update each utterance via speaker turns $S$: 
if there is an emotion-cause pair  $(U_{i},U_{j})\in L $, 
then $U_{i}=\alpha_{j,i} U_{j}+E_{i}$ ($\alpha_{j,i}\sim Unifrom ([0.7,1])$), 
and for those pairs without emotion-cause label, $\alpha_{j,i}\sim Unifrom ([0,0.3])$. 
Finally, we randomly select a noise $\xi \sim Unifrom ([-0.25,0.25])$ for each 
utterance $U_{i}=U_{i}+\xi_{i}$. 

\begin{algorithm}
  \KwInput{ $\mathcal{D},S,L$}
  \KwOutput{$SCM_{\mathcal{D}}$}
  \ForAll{$i \in {2, 3, \ldots, S}$}
  {
    \If {$Emotion(U_{i})$}
    {$E_{i} \sim \|^{50} \mathcal{N} (1,1)$}
    \Else{
      {$E_{i} \sim \|^{50} \mathcal{N} (-1,1)$}
    }
    $U_{i}\leftarrow E_{i}$
  }
  \ForAll{$i \in {1, 2, 3, \ldots, S}$}
  {  
    \ForAll{$j \in {1,2,\ldots, i}$} 
    {
      \If{$(U_{i},U_{j}) \in L$}
      {
        $U_{i}=\alpha_{j,i} U_{j}+E_{i} (\alpha_{j,i}\sim Unifrom ([0.7,1]))$\\
      }
      \Else
      {
        $U_{i}=\alpha_{j,i} U_{j}+E_{i} (\alpha_{j,i}\sim Unifrom ([0,0.3]))$\\
      }
    }   
    
  } 
  $SCM_{\mathcal{D}} \leftarrow U_{1},U_{2},\ldots, U_{S}$
  return $SCM_{\mathcal{D}}$
\caption{Creating Non-noise Synthetic dataset}
\label{aldata}
\end{algorithm}

\section{Implementation Details}\label{ID}

In the word embedding, we adopt the affect-based pre-trained features\footnote{\url{https://drive.google.com/file/d/1R5K_2PlZ3p3RFQ1Ycgmo3TgxvYBzptQG/view?usp=sharing}} 
proposed by~\citet{shen-etal-2021-directed} for all baselines and models. 

Although there are different pre-trained models in these skeleton 
baselines, the SOTA work DAG-ERC and EGAT have investigated 
their performances in a consistent pre-trained model. 
Therefore, for a fair and direct comparison, 
we continue this benchmark using the pre-trained embedding 
published by DAG-ERC for three tasks. 

In the hyper-parameters, we follow the setting of~\citet{shen-etal-2021-directed} 
in the ERC task. Moreover, in the ECPE and ECSR, the learning rate is set to 3e-5, 
batch size is set to 32, and epoch is set to 60. Further in our approach, we set $L$ 
to 1, and implicit cause size is set to 192, hidden size of GNN is set to 
300, and dropout rate is 0.3. 

Meanwhile, because there is only one training dataset for ECPE and ECSR, we 
evaluated our method ten times with different data splits by following \citet{chen2022learning} 
and then performed paired sample $t$-test on the experimental results.

Finally, we adopted downstream task modules consistent with the SOTA baselines: 
~\citet{wei-etal-2020-effective} in ECPE 
and ECSR, and ~\citet{shen-etal-2021-directed}for the ERC task.

For evaluation metrics, we follow~\citet{shen-etal-2021-directed} towards ERC, 
~\citet{xia-ding-2019-emotion} towards ECPE, 
and~\citet{poria2021recognizing} towards ECSR. Specifically, we adopt the 
macro F1 score in ECPE and ECSR tasks, micro F1 score for DailyDialog, 
and macro F1 score for the other three datasets in ERC task.

\section{Other Experiments in Affective Reasoning}\label{oear}

\begin{table}
  \footnotesize
  \centering
  \resizebox{\linewidth}{!}{
  \begin{tabular}{|c|c|c|c|c|c|}
    \hline
    Skt&Model&DailyDialog&MELD&EmoryNLP&IEMOCAP\\
    \hline
    \multirow{2}{*}{$\textbf{II}$}&DialogXL&54.93&62.41&34.73&65.94\\
    &Ours&\textbf{59.51}&\textbf{63.62}&\textbf{39.16}&\textbf{66.47}\\
    \hline
    \multirow{2}{*}{$\textbf{III}$}&EGAT$\dagger$&59.23&63.51&38.77&66.76\\
    &Ours&\textcolor{red}{\textbf{59.68}}&\textbf{63.71}&\textbf{39.62}&\textbf{68.18}\\
    \hline
    \multirow{2}{*}{$\textbf{IV}$}&RGAT&54.31&60.91&34.42&65.22\\
    &Ours&\textbf{59.65}&\textbf{63.69}&\textbf{39.22}&\textbf{67.65}\\
    \hline
    \multirow{2}{*}{$\textbf{V}$}&DECN$\dagger$&59.08&63.78&39.44&67.41\\
    &Ours&\textbf{59.28}&\textcolor{red}{\textbf{63.91}}&\textcolor{red}{\textbf{40.11}}&\textbf{67.61}\\
    \hline
    \multirow{2}{*}{$\textbf{VI}$}&DAG-ERC&59.33&63.65&39.02&68.03\\
    &Ours&\textbf{59.53}&\textbf{63.81}&\textbf{39.54}&\textcolor{red}{\textbf{69.17}}\\
    \hline
  \end{tabular}}
  \caption{Overall performance in ERC task. $\dagger$ denotes the 
  results implemented in this paper. 
  The better scores in the same skeleton are in bold, and the best of all skeletons is in red.}
  \label{taberc}
\end{table}

In Table~\ref{taberc}, our approach performs better than the corresponding 
baseline under all skeletons in four datasets. Hence, using a causal 
auto-encoder to find the implicit causes benefits this task. 
Besides, our approach improves significantly under skeletons 
$\textbf{II}$, $\textbf{III}$, and $\textbf{IV}$. From 
Figure~\ref{figskeleton}, these three skeletons have more relevant 
nodes than others, so there are more redundant edges 
to be corrected by our approach, which is demonstrated again in Appendix E. 
In contrast, $\textbf{V}$ and $\textbf{VI}$ achieve the best results in 
MELD, EmoryNLP, and IEMOCAP datasets, 
which indicates that \textbf{Hypothesis 5} is more probably a strong inductive 
bias that conversation enjoys. 

\begin{figure}
  \includegraphics[width=1\linewidth]{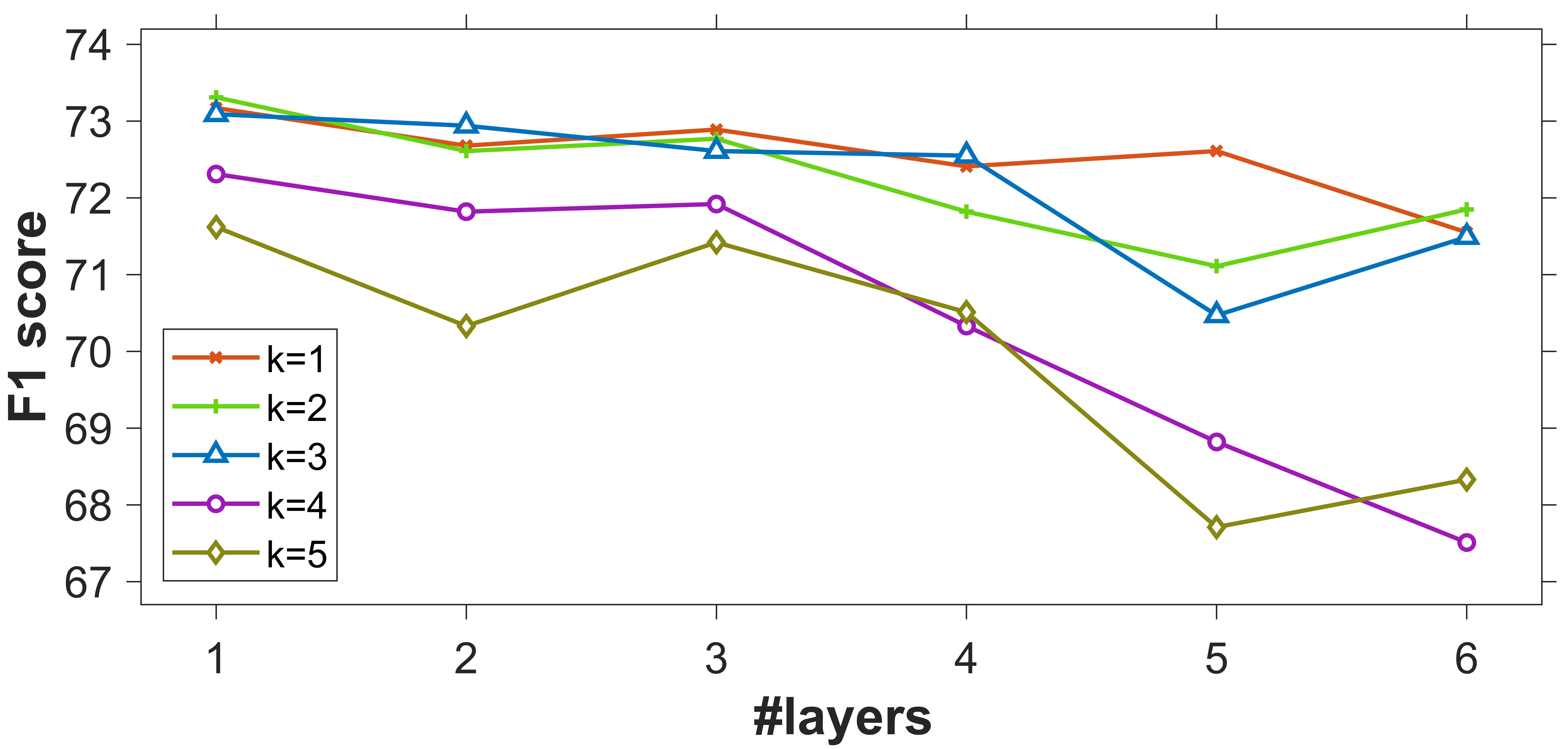}
  \caption{Further layers $L$ and related node number $K$ 
  with \textbf{VI} skeleton model in ECPE task. }
  \label{figlayers}
\end{figure}
Then, we investigate how the number of layers and 
the variants of causal skeletons would affect the performance of 
our approach. So we further conducted several contrasts 
with $k$ up to 5 and $L$ up to 6, as shown in Figure~\ref{figlayers}. 
One observation is that the best 
performance occurs at either $k=1$, $2$, or $3$, which indicates that 
$k\geqslant 4$ offers no advantage and even leads to confounding. Moreover, 
$L=1$ achieves the best performance under all $k$ values. In other words, 
one layer is sufficient to yield the most effective implicit causes.

\section{Visualization of Causal Graph}\label{vcg}
\begin{figure*}[hbp]
  \includegraphics[width=0.95\linewidth]{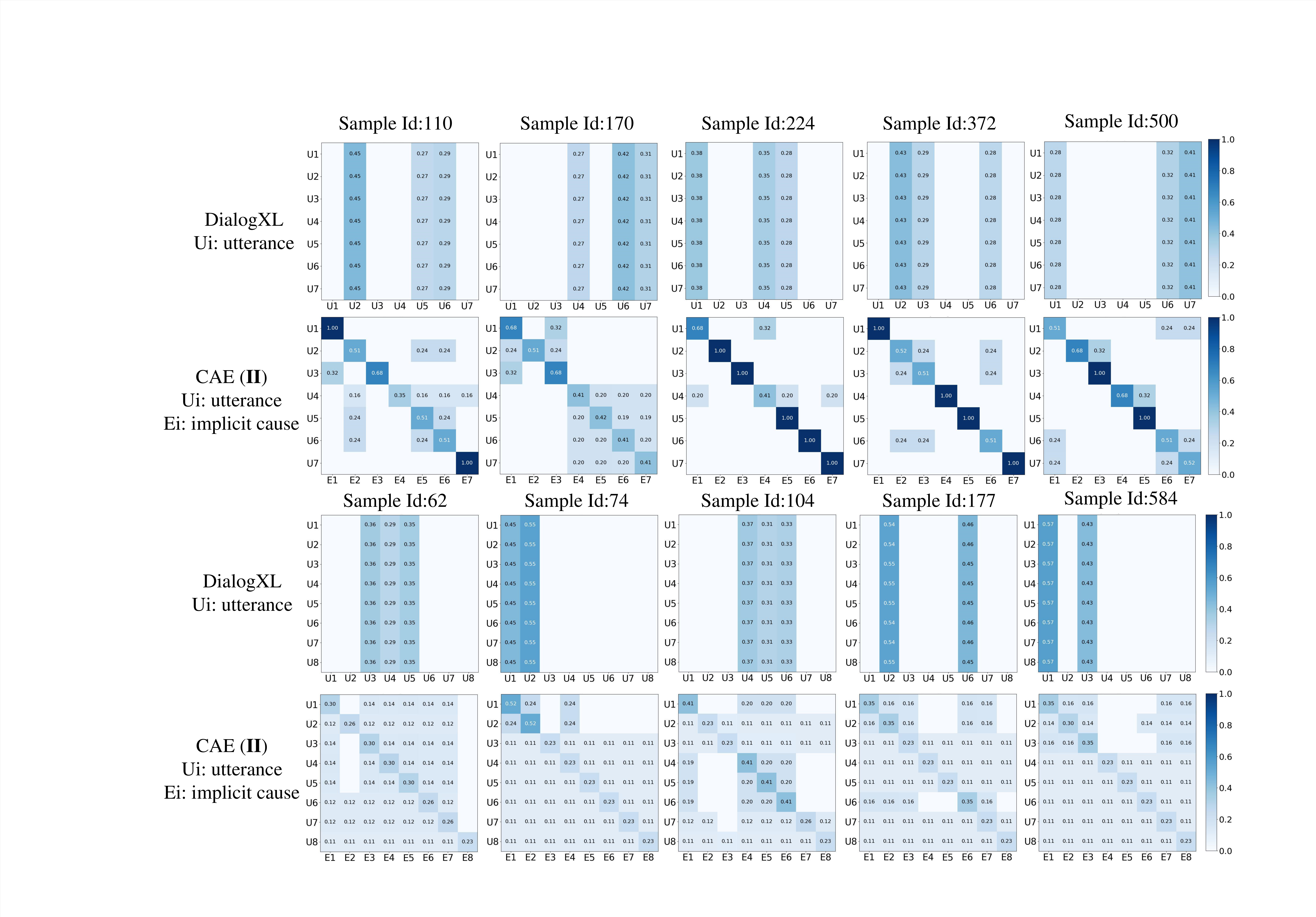}
  \caption{Causal Graph cases of DialogXL and Ours (CAE $\textbf{II}$). }
  \label{figskeleton_case2}
\end{figure*}

\begin{figure*}
  \includegraphics[width=0.95\linewidth]{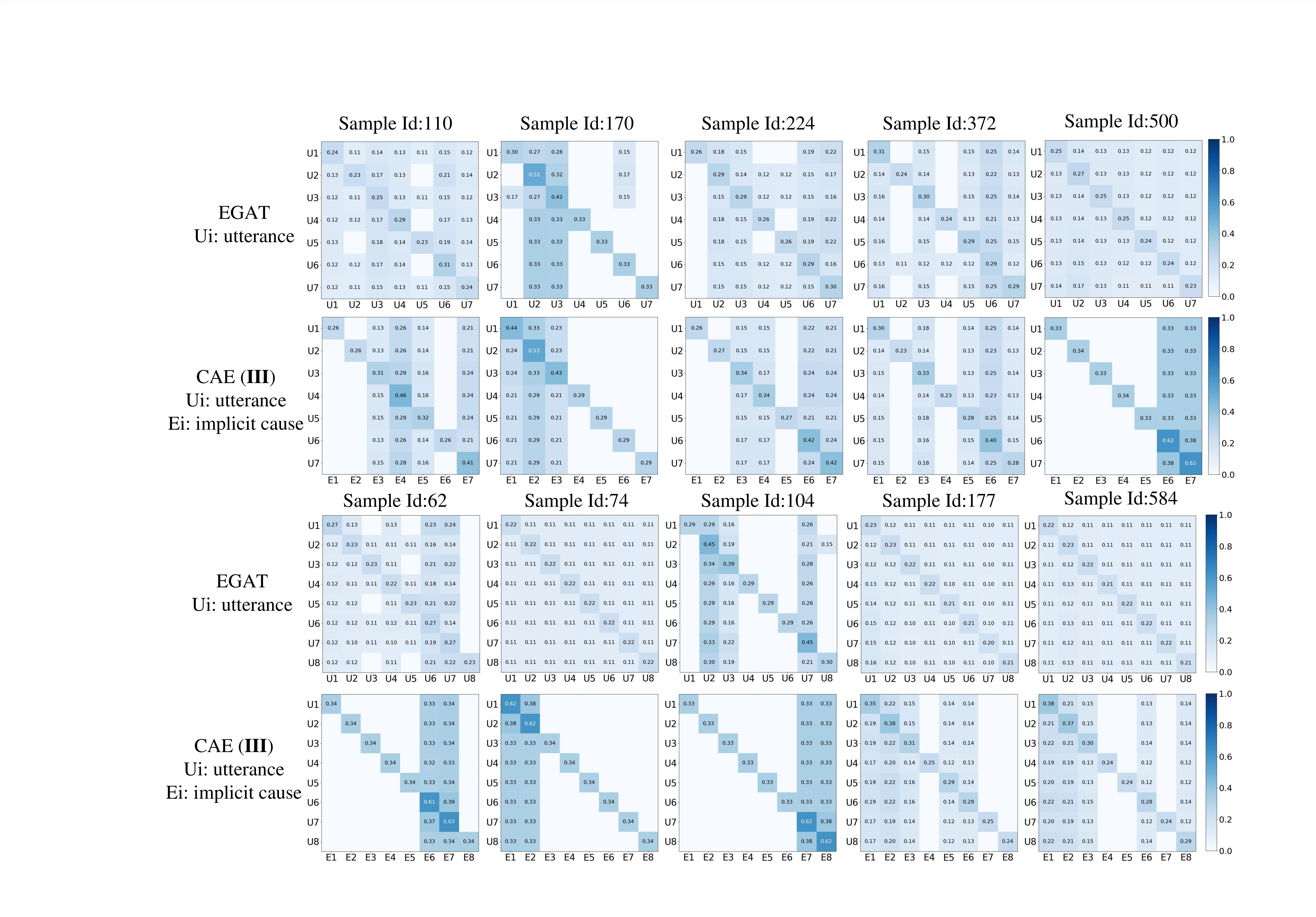}
  \caption{Causal Graph cases of EGAT and Ours (CAE $\textbf{III}$). }
  \label{figskeleton_case3}
\end{figure*}

\begin{figure*}
  \includegraphics[width=0.95\linewidth]{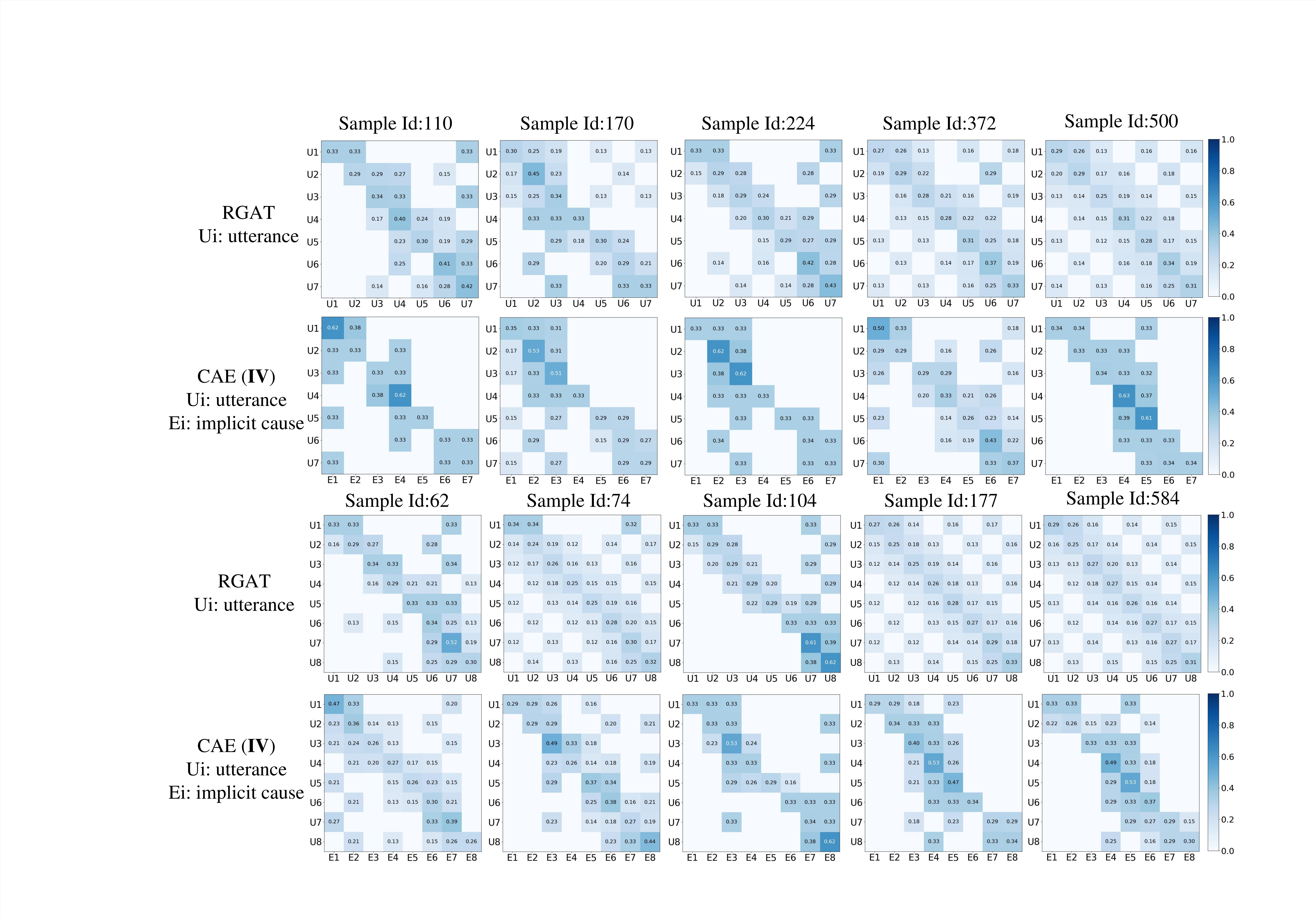}
  \caption{Causal Graph cases of RGAT and Ours (CAE $\textbf{IV}$). }
  \label{figskeleton_case4}
\end{figure*}

\begin{figure*}
  \includegraphics[width=0.95\linewidth]{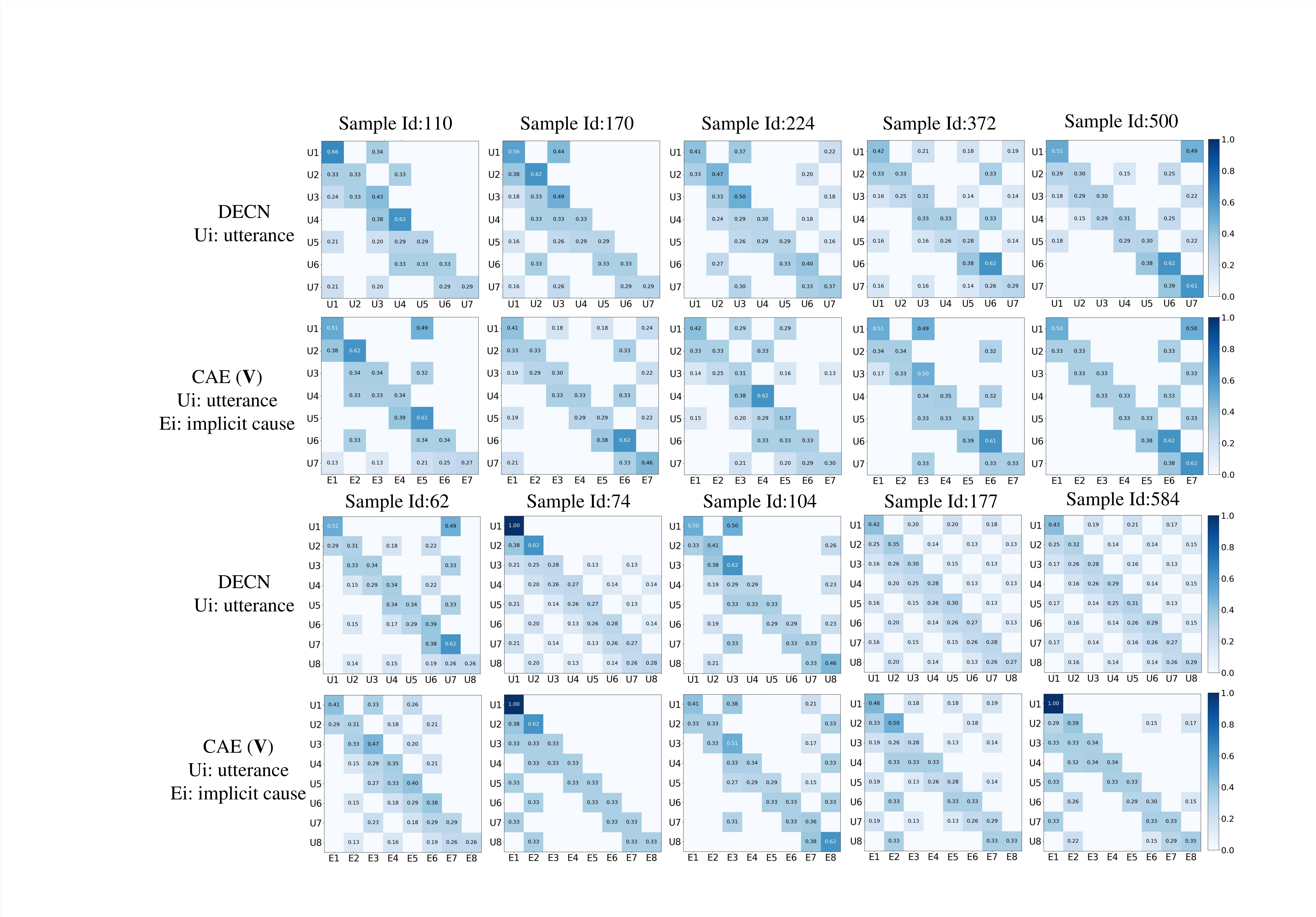}
  \caption{Causal Graph cases of DECN and Ours (CAE $\textbf{V}$). }
  \label{figskeleton_case5}
\end{figure*}

\begin{figure*}
  \includegraphics[width=0.95\linewidth]{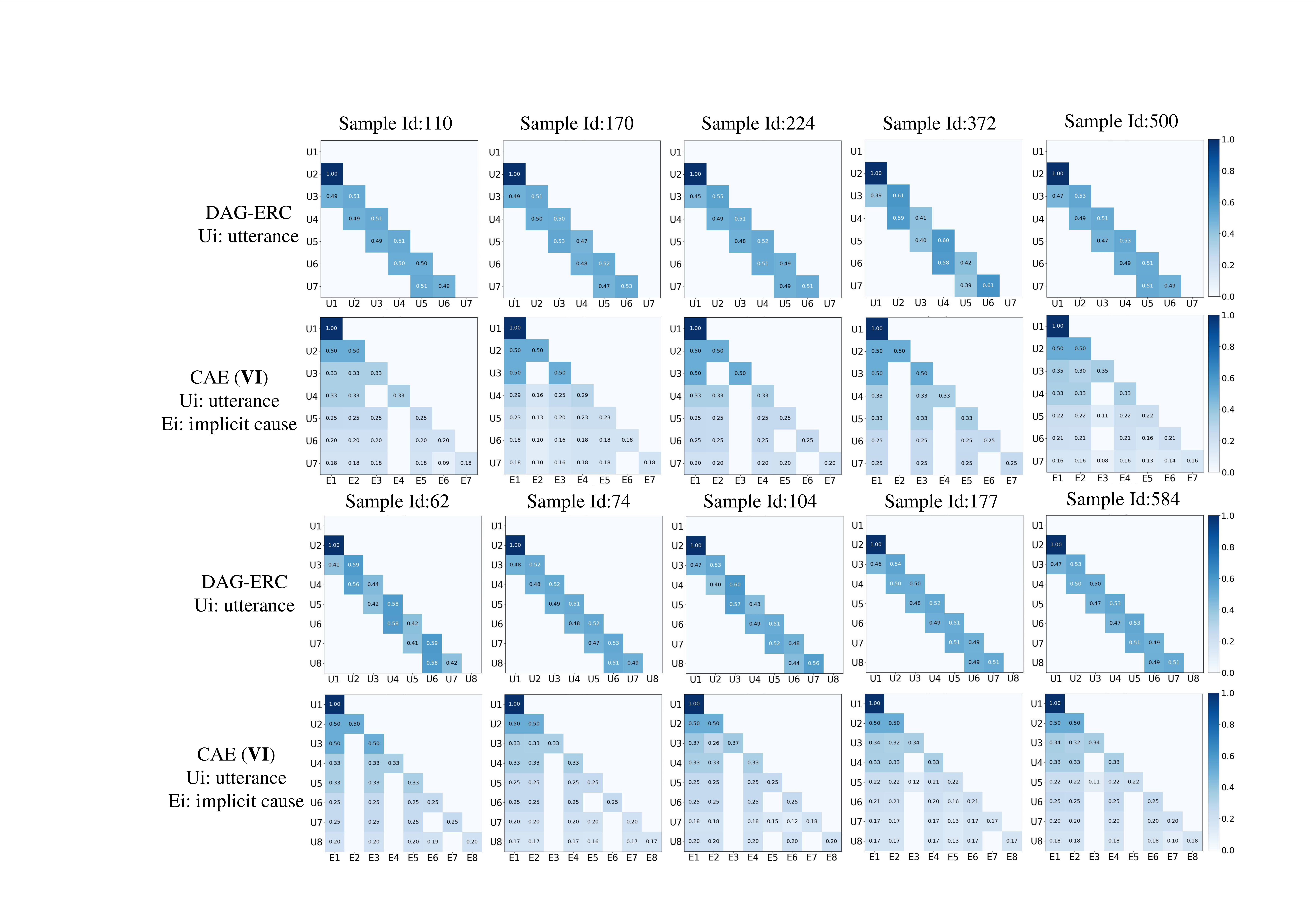}
  \caption{Causal Graph cases of DAG-ERC and Ours (CAE $\textbf{VI}$). }
  \label{figskeleton_case6}
\end{figure*}

In the Figure~\ref{figskeleton_case2} to ~\ref{figskeleton_case6}, 
we showed the Visualization of the adjacency matrix $(I-A^{T})^{-1}$. When the 
auxiliary loss $Loss_{KL}$ achieves the lower bound, $(I-A^{T})^{-1}$ represents the 
relationship matrix between utterances and implicit causes. 

In the ECPE task, we extracted 10 samples from test sets in different 
folds. To facilitate comparison and contrasting, we selected five  
7-utterances cases and five 8-utterances cases. The IDs are as following: 

\textbf{7-utterances cases}: 110, 170, 224, 372, 500. 

\textbf{8-utterances cases}: 62, 74, 104, 177, 584.

\renewcommand{\dblfloatpagefraction}{0.9}

To obtain the non-negative value, we adopted the $T=sigmoid(\cdot)-0.05$ to process 
the original tensors $(I-A^{T})^{-1}$ outputted from the encoder. We follow a 
common practice: set the threshold as $0.05$ to delete some unimportant edges. 
And to highlight which implicit cause contributes the each utterance best, 
we adopted the $sofmax(\cdot)$ to process columns afterward and labeled 
the block with value $>0$.

It is excepted that: (i) when skeletons construct overage edges, our 
model is able to degrade the influences of some negligible utterances 
by deleting the corresponding edges from their implicit causes. (ii) when 
skeletons construct insufficient edges, our model can add some edges 
to obtain more information.

\section{Proof of emotion consistency of implicit causes and utterances}
\label{pecicu}

\begin{table}
  \footnotesize
  \centering
  \resizebox{\linewidth}{!}{
  \begin{tabular}{|c|c|c|c|c|}
    \hline
    Skt&DailyDialog&MELD&EmoryNLP&IEMOCAP\\
    \hline
    
    $\textbf{II}$&51.48 ($\downarrow$8.03)&\textbf{58.41 ($\downarrow$5.21)}&34.97 ($\downarrow$4.19)&59.71 ($\downarrow$6.76)\\
    \hline
    
    $\textbf{III}$&54.37 ($\downarrow$5.31)&58.19 ($\downarrow$5.52)&36.55 ($\downarrow$3.07)&63.42 ($\downarrow$4.76)\\
    \hline
    
    $\textbf{IV}$&\textbf{55.62 ($\downarrow$4.03)}&57.22 ($\downarrow$6.47)&\textbf{36.91 ($\downarrow$2.31)}&62.34 ($\downarrow$5.31)\\
    \hline
    
    $\textbf{V}$&54.62 ($\downarrow$4.66)&58.19 ($\downarrow$5.72)&35.49 ($\downarrow$4.62)&63.13 ($\downarrow$4.48)\\
    \hline
    
    $\textbf{VI}$&53.27 ($\downarrow$6.26)&58.39 ($\downarrow$5.42)&34.98 ($\downarrow$4.56)&\textbf{65.18 ($\downarrow$3.99)}\\
    \hline
  \end{tabular}}
  \caption{Overall performance of implicit causes $E$ in ERC task.}
  \label{tablatenterc}
\end{table}

We would like to explain why implicit causes and utterances are consistent 
in emotion from both theory and euqation, in the condition where 
emotional utterance and cause utterance possess different emotion types. 

We define the implicit causes as the unobservable emotional desire 
and the utterances as the observable emotional expression. 
This definition is proposed in
~\citet{ong2019computational,ong2015affective}, 
which also argues that emotional expression is affected by desires 
and event outcomes. Moreover, for emotion utterances that 
are not influenced by explicit cause factors, 
the source of their emotions should originate from implicit causes. 
The desire and the expression generally belong to the same emotion 
because the outcomes often have little effect on emotional 
expression. 
Our paper can also deduce this conclusion from the SCM (Equation~\ref{eqn1}). 
Considering there is a linear map $f(\cdot)$ from representation 
space to emotion space. Then we can obtain the following: 
\begin{equation}
 f((I-A) U)=f(E)
 \label{eqn12}
\end{equation}
\begin{equation}
 (I-A) f(U)=f(E)
 \label{eqn13}
\end{equation}
\begin{equation}
 f(U)=W^{T}f(E)
 \label{eqn14}
\end{equation}
Note that $W=(I-A)$ and $A_{i,i}=0$. So in $W$, 
the value of the elements on the diagonal is constant at $1$ and 
is a constant maximum of each column. 
Naturally, $f(E)$ is an approximate estimate of $f(U)$ especially 
$U_{t}$ and $U_{i}$ in the ECP $(U_{t}, U_{i})$ do not belong to the 
same emotion category, 
which is why we think implicit causes are reasonable when the F1 
score of Table 6 is high. 

Therefore, we test the F1 scores in ERC task 
by replacing $\widehat{H}$ with $E$ from a consensus that 
implicit causes should be aligned with utterances in the emotion types. 

In Table~\ref{tablatenterc}, we reported the overall results 
of $E$ in ERC task. Note that we only examine the sample of 
ECP with different emotion types. 
Among five skeletons and four datasets, almost all results 
achieve $90\%$ scores of corresponding performances of $\widehat{H}$, 
which indicates that $E$ is practically aligned with 
$\widehat{H}$ in the affective dimension.

\end{document}